%% file: bare_jrnl_new_sample4.tex
\documentclass[lettersize,journal]{IEEEtran}
\usepackage{amsmath,amsfonts}
\usepackage{algorithmic}
\usepackage{algorithm}
\usepackage{array}
\usepackage[caption=false,font=normalsize,labelfont=sf,textfont=sf]{subfig}
\usepackage{textcomp}
\usepackage{stfloats}
\usepackage{url}
\usepackage{verbatim}
\usepackage{graphicx}
\usepackage{cite}
\usepackage{booktabs}
\usepackage{multirow}
\usepackage[table,xcdraw]{xcolor}
\usepackage[normalem]{ulem}
\useunder{\uline}{\ul}{}

\begin{document}

\title{SIDNet: Learning Shading-aware Illumination Descriptor for Image Harmonization}

\author{Zhongyun Hu, Ntumba Elie Nsampi, Xue Wang, and Qing Wang,~\IEEEmembership{Senior~Member,~IEEE}
        % <-this % stops a space
\thanks{Z. Hu, N. Nsampi, X. Wang and Q. Wang (corresponding author) are with the School of Computer Science, Northwestern Polytechnical University, Xi’an 710072, China (E-mail: qwang@nwpu.edu.cn).} %The work was supported by NSFC under Grant 62031023.}% <-this % stops a space
\thanks{Z. Hu and N. Nsampi contributed equally to this work.}
\thanks{Project website: https://waldenlakes.github.io/IllumHarmony/}
\thanks{Manuscript received April 19, 2021; revised August 16, 2021.}}

\markboth{Journal of \LaTeX\ Class Files,~Vol.~14, No.~8, August~2021}%
{Shell \MakeLowercase{\textit{et al.}}: A Sample Article Using IEEEtran.cls for IEEE Journals}


\maketitle

\begin{abstract}
   Image harmonization aims at adjusting the appearance of the foreground to make it more compatible with the background. Without exploring background illumination and its effects on the foreground elements, existing works are incapable of generating a realistic foreground shading. In this paper, we decompose the image harmonization task into two sub-problems: 1) illumination estimation of the background image and 2) re-rendering of foreground objects under background illumination. Before solving these two sub-problems, we first learn a shading-aware illumination descriptor via a well-designed neural rendering framework, of which the key is a shading bases module that generates multiple shading bases from the foreground image. Then we design a background illumination estimation module to extract the illumination descriptor from the background. Finally, the Shading-aware Illumination Descriptor is used in conjunction with the neural rendering framework (SIDNet) to produce the harmonized foreground image containing a novel harmonized shading. Moreover, we construct a photo-realistic synthetic image harmonization dataset that contains numerous shading variations with image-based lighting. Extensive experiments on both synthetic and real data demonstrate the superiority of the proposed method, especially in dealing with foreground shadings.
\end{abstract}


\section{Introduction}

\begin{figure*}[ht]
    \centering
    \includegraphics[width=0.95\linewidth]{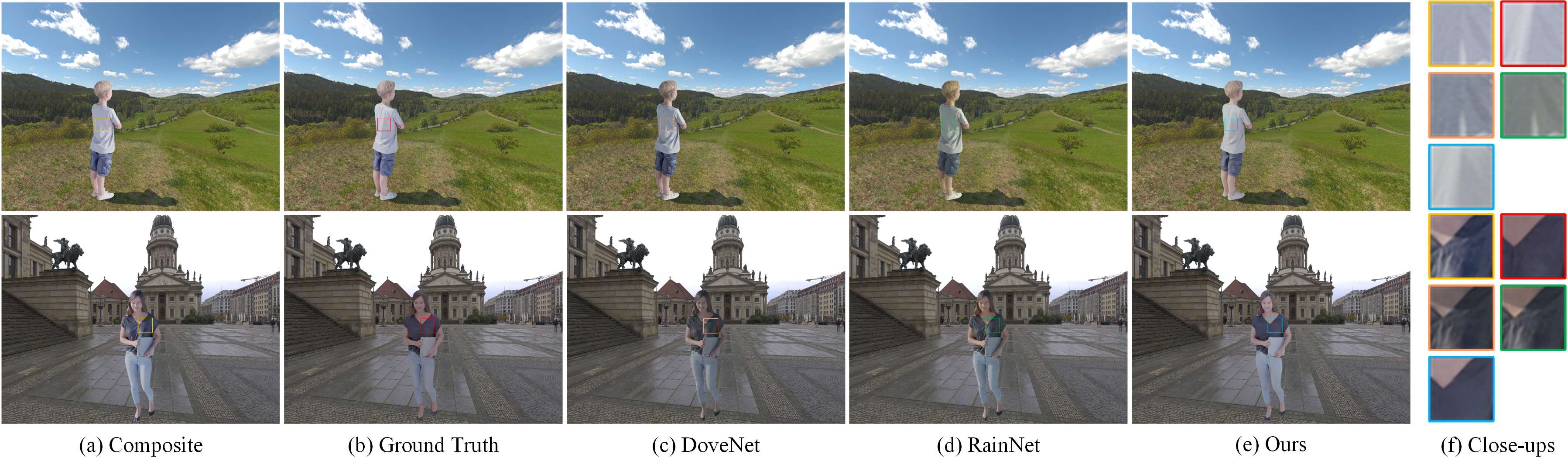}
    \caption{Given a composite image (a) of which the foreground and background taken from different images, our proposed method is able to produce the harmonized image (e) containing a novel foreground shading (f) that conforms to the background illumination. In contrast, existing methods \cite{Cong2020DoveNet, ling2021region} only adjust the brightness and color of the foreground. Here, for both methods, we use their publicly released pre-trained models.}
    \label{fig:teaser}
\end{figure*}

\IEEEPARstart{G}{iven} a composite image of which the foreground and background taken from different images, image harmonization aims to adjust the appearance of the foreground to make it compatible with the background. A lot of works \cite{lalonde2007using,xue2012understanding,zhu2015learning,tsai2017deep,Cong2020DoveNet,cong2021deep,guo2021iih,ling2021region} have been proposed to solve the inharmony problem in the composite image. As shown in Fig.~\ref{fig:teaser}, these image harmonization methods, however, tend to focus on adjusting the low-level statistics (i.e., color and brightness) of the foreground rather than its shading.

The failure to model shadings can be attributed to the lack of a comprehensive understanding of the background illumination and its effects on the foreground elements, especially the direction and distribution of illumination cannot be perceived by these methods. In particular, learning-based approaches, such as \cite{tsai2017deep,Cong2020DoveNet,guo2021iih,ling2021region}, are generally formulated as the image-to-image translation task where the illumination is implicitly transferred from the background to the foreground. Moreover, existing large-scale image harmonization datasets \cite{tsai2017deep,Cong2020DoveNet} are devoid of perceivable shading variations. It is questionable whether or not the networks trained with these datasets could deal with shading variations.

Collecting, processing, and distributing real-world datasets is often associated with data gathering costs, quality problems, and privacy concerns. Recently, researchers have turned to synthetic data as an adequate solution in the face of the numerous data-related challenges posed by real-world data and the requirements of recent AI technologies \cite{Cong2020DoveNet}. Several color transfer algorithms \cite{reinhard2001color,xiao2006color,pitie2007automated,fecker2008histogram} have been used to generate visually acceptable images with varying colors and brightness. However, the shading variation, which is also important for image harmonization, has not been taken into consideration. To this end, we construct a large-scale photo-realistic image harmonization dataset that contains color, brightness and shading variations with image-based lighting \cite{debevec2006image}. Unlike existing synthetic datasets \cite{cong2021deep, bao2022deep} of which the foreground objects or the illumination maps are created by CG software, we refer to real models/illumination captured from the real world, with the aim of achieving photo-realistic renderings.

In this paper, we propose to decompose the image harmonization task into two sub-problems: (1) illumination estimation of the background image, and (2) re-rendering of foreground objects under background illumination.
The general lighting representation (i.e. illumination maps \cite{Debevec1997HDR}) that is able to record the complete illumination (including the directional information) can be used to solve sub-problem (1). However, using such a representation brings considerable challenges due to its large number of parameters. Our key to solving (1) lies in the proposal of an efficient illumination representation with fewer parameters that also can retain directional information. For sub-problem (2), in contrast to the spherical harmonics lighting model \cite{ramamoorthi2001efficient}, we intend to render complex global illumination effects such as cast shadows to further improve the realism of the composite image. 

To achieve these ends, we propose a novel Neural Rendering Framework that accounts for global illumination effects while learning a shading-aware illumination descriptor from the illumination maps. 
Its key component is a neural Shading Bases Module, which is utilized to generate multiple shading bases from the foreground image. Each shading basis corresponds to a specific illumination distribution. It then combines with the illumination descriptor, which is encoded by an Illumination Encoder Module, to render a shading image. We propose to reconstruct the shading image as a pretext task in order to simultaneously supervise the learning of both the shading bases and the illumination descriptor. Note that the GT shading image is automatically generated by a path tracing algorithm, and also contains global illumination effects.

Once we pre-define the shading-aware illumination descriptor, the illumination of the background image could be estimated via the proposed Background Illumination Estimation Module and then is used in conjunction with our Neural Rendering Framework to generate the harmonized foreground image which contains a harmonized shading. We name this novel Shading-aware Illumination Descriptor-based image harmonization Network SIDNet.

Our contributions can be summarized as follows:

(1) We propose a neural Shading Bases Module, which decomposes the shading field into multiple shading components, to generate a novel foreground shading using the estimated illumination descriptor. To the best of our knowledge, this is the first of its kind to explicitly model shadings in image harmonization.

(2) We design a novel Neural Rendering Framework to learn the shading-aware illumination descriptor from the illumination maps in a self-supervised manner.

(3) We provide a large-scale photo-realistic synthesized image harmonization dataset containing challenging shading variations.

\section{Related Work}
In this section, we briefly discuss existing works related to image harmonization. In addition, image relighting methods relevant to the proposed work are also included.

\subsection{Image Harmonization}
Traditional image harmonization methods \cite{reinhard2001color, perez2003poisson, pitie2005ndimensional, cohen2006color, Jia2006Pasting, pitie2007automated, lalonde2007using, sunkavalli2010multi, xue2012understanding} focus on matching low-level statistics between images. The pioneer work \cite{reinhard2001color} matched the means and variances of the color histograms between images in a decorrelated color space. Lalonde and Efros \cite{lalonde2007using} then combined global color statistics obtained over a large natural image set and local color statistics to improve the realism of the composite images. Sunkavalli et al. \cite{sunkavalli2010multi} proposed to match contrast, texture, noise, and blur of visual appearance using multi-scale pyramid representations to produce realistic composites. Xue et al. \cite{xue2012understanding} identified key statistical measures that most affected the realism of a composite image. However, the adjustment of low-level statistics can not handle shading variations.

In the past few years, researchers have concentrated on deep neural network-based approaches for image harmonization \cite{zhu2015learning, tsai2017deep, song2020illumination, Cong2020DoveNet, guo2021iih, ling2021region, sofiiuk2021foreground, jiang2021ssh, cong2022high}. Zhu et al. \cite{zhu2015learning} proposed a CNN-based classifier model for the perception of realism to guide a traditional color adjustment method to produce more realistic outputs. The first end-to-end CNN model for image harmonization was proposed by Tsai et al. \cite{tsai2017deep}. These methods usually formulate image harmonization as an image-to-image translation task by ensuring visual consistency between the foreground and the background in different aspects, such as the domain consistency \cite{Cong2020DoveNet,cong2021bargainnet}, the visual style consistency \cite{ling2021region, hang2022scs}, and the reflectance/illumination consistency \cite{guo2021iih, bao2022deep}. 
Furthermore, the attention \cite{cun2020improving, hao2020image} or self-attention \cite{guo2021dht} mechanism is applied to improve the realism of the composite images. However, without considering the physical principles of image formation, learning-based methods lack the perception of illumination information in the background image, especially the direction of illumination. This inevitably leads to their inability to generate a realistic foreground shading, which severely degrades the realism of the composite images. 
%In contrast, we propose to extract a direction-aware descriptor from the background image, and then apply it in a neural rendering framework to generate a new foreground shading that conforms to the background illumination.

\subsection{Image Relighting}
Traditional image-based relighting methods \cite{debevec2000acquiring,xu2018deep,meka2019deep} directly reconstruct the light transport function to relight the objects using multiple images under different illumination conditions. Note that the illumination here is always explicitly provided. Recently, several deep neural networks with illumination estimation modules \cite{zhou2019deep,sun2019single,kanamori2018relighting,wang2020single,sang2020single, lagunas2021single} are proposed to relight objects of a specific class (e.g., portraits and human bodies) using a single RGB image. Still, the illumination estimation is only considered for specific objects rather than the natural scenes. Given multi-view images, Yu et al. \cite{yu2020self} proposed the first single image-based outdoor scene relighting method along with lighting estimation for the scene. They used spherical harmonics lighting model \cite{ramamoorthi2001efficient} to generate the shading. However, it could not handle global illumination. Moreover, it is worth noting that although these relighting methods \cite{yu2020self,zhou2019deep,sun2019single,kanamori2018relighting,wang2020single,sang2020single, lagunas2021single} with illumination estimation can be applied to image harmonization, additional computational overhead would be also introduced, since illumination estimation for the background image is often accompanied by meanwhile estimating other physical attributes in the background image. In other words, these relighting methods are not specifically designed for image harmonization. % Differently, we consider the problem of image harmonization using a single RGB image, hence how to utilize the depth information is the key to our method.

\begin{figure*}[ht!]
    \centering
    \includegraphics[width=0.9\linewidth]{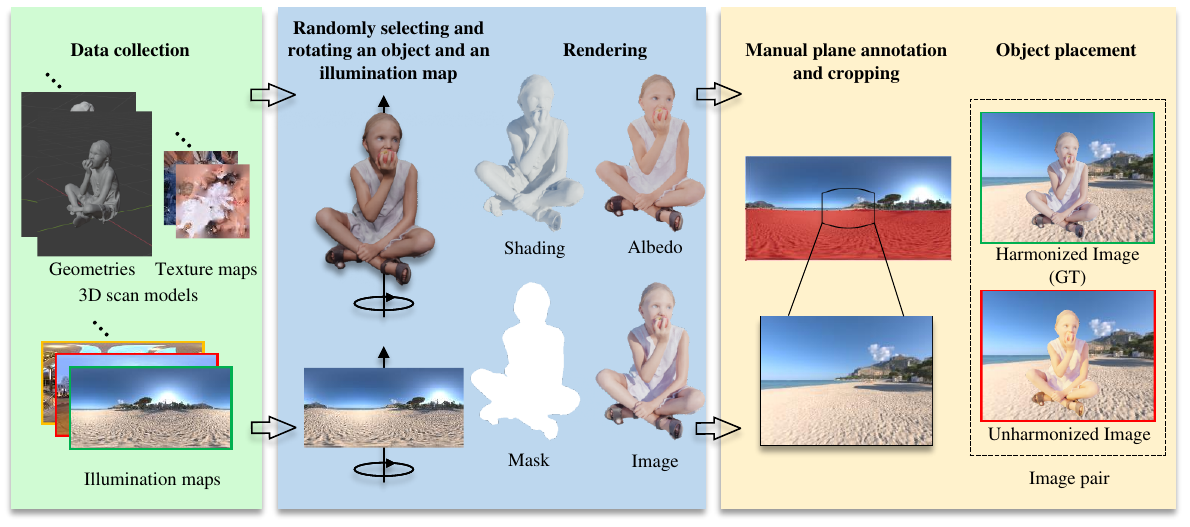}
    \caption{The pipeline of constructing the pair of harmonized image and unharmonized image. It mainly covers data collection, rendering, and object placement. To train the proposed model, shading and albedo are also rendered.}
    \label{fig:dataset construction pipeline}
\end{figure*}

\section{Image Harmonization Dataset}
The goal of our image harmonization dataset is to introduce a new challenging benchmark with photo-realistic synthesized images and plentiful variations (color, brightness, and shading) to the community of image harmonization. The pipeline of constructing the pair of harmonized image (ground truth) and unharmonized image is illustrated in Fig.~\ref{fig:dataset construction pipeline}. Below we introduce the construction process, which covers data collection, rendering and object placement.

\subsection{Data Collection}
To construct our dataset, we collect both high-quality 3D human models and high dynamic range (HDR) illumination maps. The collected 3D models are acquired from \cite{3dpeople} using photogrammetric 3D scanning methods. A rich variety of humans are included, with the diversity across genders (male, female), ages, poses, and clothing (colors, accessories). We collect a total of 138 high-quality 3D humans, of which 120 are used for training and 18 for testing.

\begin{figure}
    \centering
    \includegraphics[width=0.99\linewidth]{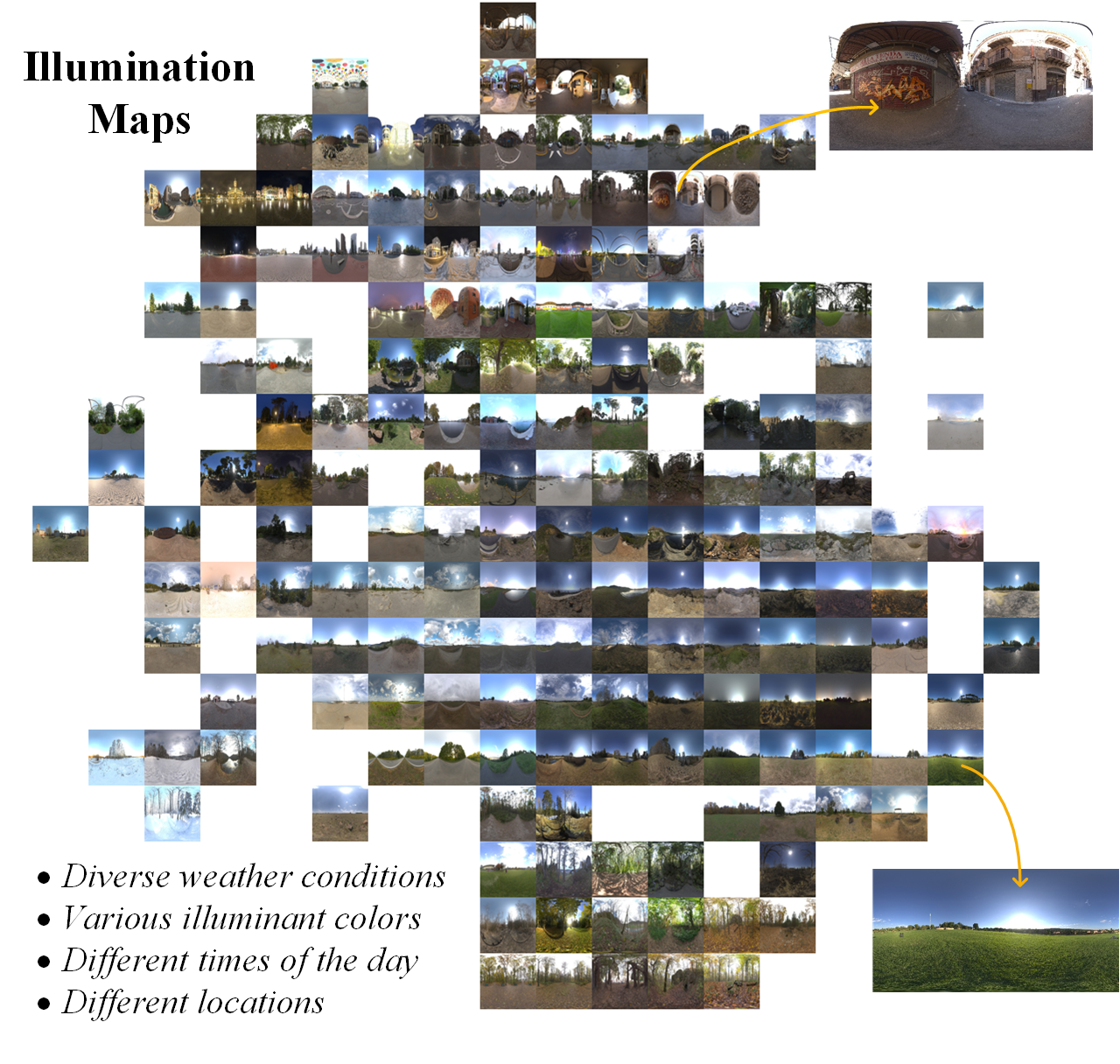}
    \caption{The t-SNE visualization of our collected illumination maps, which contain diverse weather conditions, illuminant colors, time of the day, and locations. }
    \label{fig:tSNE_visualization}
\end{figure}

Our illumination maps are collected from the internet source Poly Haven \cite{polyHaven} and HDR MAPS \cite{hdrmaps}, which offer diverse high dynamic range panoramic images. Fig.~\ref{fig:tSNE_visualization} shows the t-SNE visualization of our illumination maps. We mainly select outdoor illumination maps, resulting in a total of 318 high dynamic range panoramic images. In order to generate images with different kinds of variations (color, brightness, and shading), we ensure that the selected illumination maps are diverse across weather conditions (sunny, cloudy, and overcast), illuminant colors, time of the day, and locations. From the 318 images, 191 images are used for training and the remaining 127 ones are reserved for testing. All the selected illumination maps come with the resolution of 8k, which are resized to 2k before rendering.

\subsection{Rendering}
\begin{figure}
    \centering
    \includegraphics[width=0.99\linewidth]{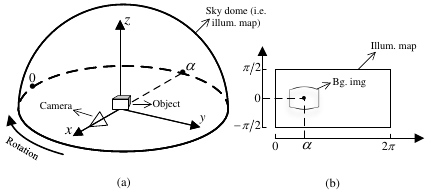}
    \caption{The Blender rendering system (a) and the cropping horizontal center $\alpha$ of the illumination map for the background image (b).}
    \label{fig:blender_rendering_system}
\end{figure}

To generate training and test images, we use Blender\cite{blender} with Cycle Renderer. Each object is first placed on a planar surface within Blender's environment. We then randomly sample (without replacement) half of the images as illumination maps. For each possible pair of object and illumination map, we randomly sample 4 rotation angles from a pre-defined set of 8 angles, ranging from 0 to 360 degrees with an increment of 45 degrees. The sampled angles are used to rotate both the object and the illumination map, resulting in a total of 16 path-traced images per object. This process increases the richness of object poses and provides sufficient shading variations for our model to learn. Specifically, as shown in Fig.~\ref{fig:blender_rendering_system}(a), a Blender rendering system includes three parts: a sky dome, a camera, and an object. We randomly select one illumination map as the sky dome. In order to generate various shading variations on the foreground object, we need to first rotate the illumination map by a randomly sampled angle, and then render a foreground image. Assume that the horizontal coordinate of the rotated illumination map pointed by the camera is $\alpha$ at this time. As shown in Fig.~\ref{fig:blender_rendering_system}(b), the illumination map must be cropped with $\alpha$ as the horizontal center to obtain the background image to ensure the illumination consistency between the rendered foreground image and the background image. In the next subsection, this illumination consistency enables the foreground image to be placed within the background image.

For each object, we generate path-traced images, shading images, albedo images, and foreground masks, which are all rendered with 480 $\times$ 640 resolution. About 200$\sim $300 samples per pixel are used for generating path-traced images.

\subsection{Object Placement}\label{obj_placement}

The location of an object within an image conveys important clues for image harmonization. Here, object placement and tuple building for training and test sets will be elaborated. We assume that all objects are placed on planar surfaces. To distinguish between planar and non-planar surfaces within an image, we manually annotate the planar surface in the illumination map from which the background image will be extracted.

For a given background image extracted from an illumination map using a virtual perspective camera, we randomly select a pixel belonging to the annotated planar surface. Note that we discard the background images that do not contain the annotated plane surface.
%Depending on the location of the selected pixel within the cropped image
We then crop the rendered object as the foreground image and select one image corner as the reference point. Finally, we randomly resize the cropped object and compose it with the background image, so that the reference point uses the randomly selected pixel.

\begin{figure}
    \centering
    \includegraphics[width=\linewidth]{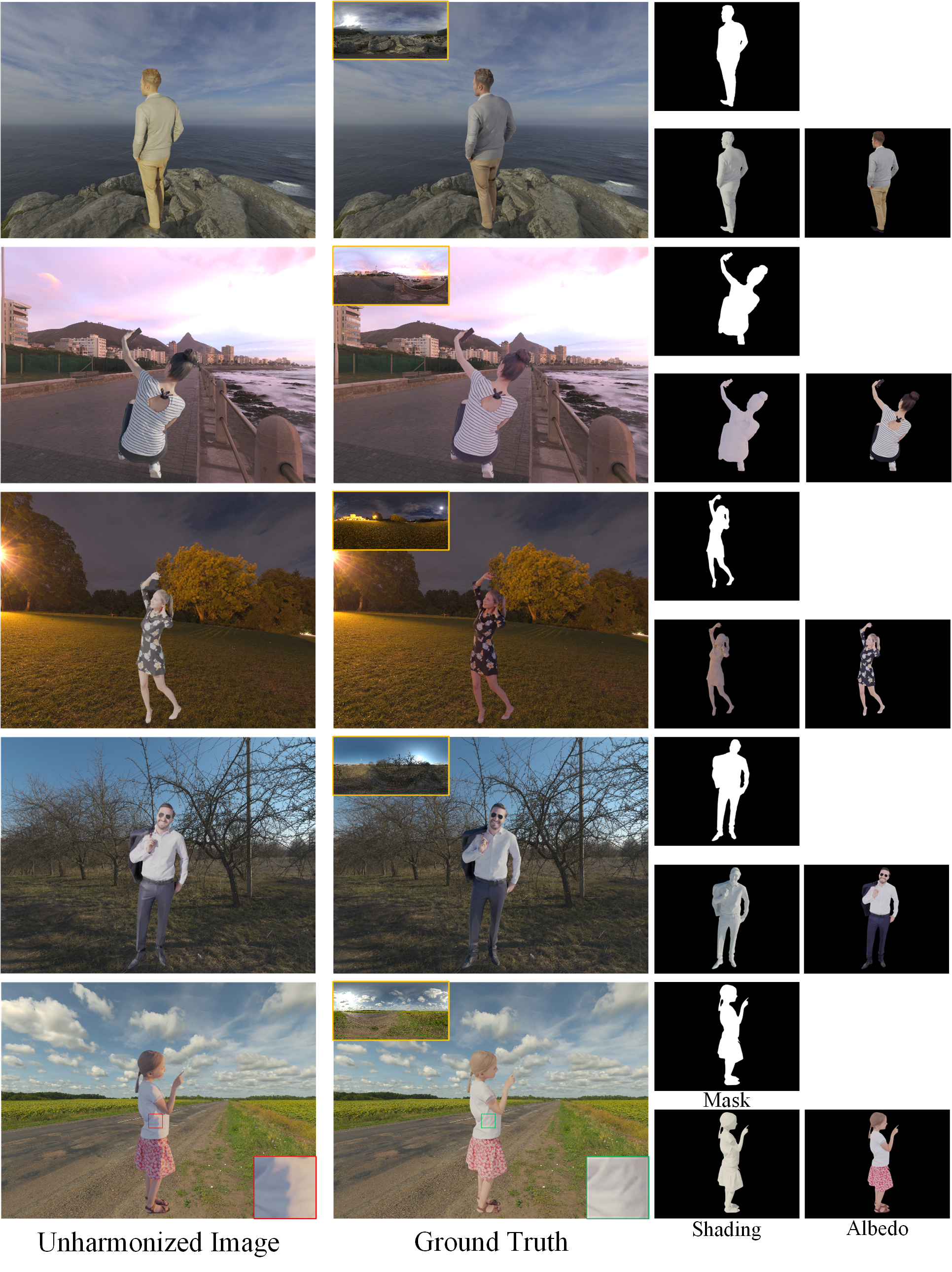}
    \caption{High-quality examples from our constructed dataset. Red and green insets in the bottom row indicate that our dataset contains challenging \emph{shading variations}.}
    \label{fig:examples_from_dataset}
\end{figure}

To create the training and test tuple, we first select a rendered image of an object and its corresponding illumination map. We then rotate the illumination map based on the angle used for rendering and extract an image crop (background image) using a virtual perspective camera. Here, a standard gamma tone mapping ($\gamma = 2.2$ ) is also applied to the illumination map before extracting the background image. Lastly, we perform object placement and compose foreground/background images as described above. The same procedure is used to create unharmonized and harmonized images, only the unharmonized image contains the same object rendered under a different illumination. Fig.~\ref{fig:examples_from_dataset} shows some representative examples from our constructed dataset.

\subsection{Dataset Summary}
Our dataset has a total of 143,390 training images and 22,048 test images, which cover a wide range of scenes and illumination conditions. We further split the training set and the test set into four categories based on illumination conditions as reported in Tab.~\ref{tab:dataset statistics}. The binary foreground (object) mask is also provided for each image.

\input{Tables/dataset_statistics}

\section{Method}
\begin{figure*}
    \centering
    \includegraphics[width=\linewidth]{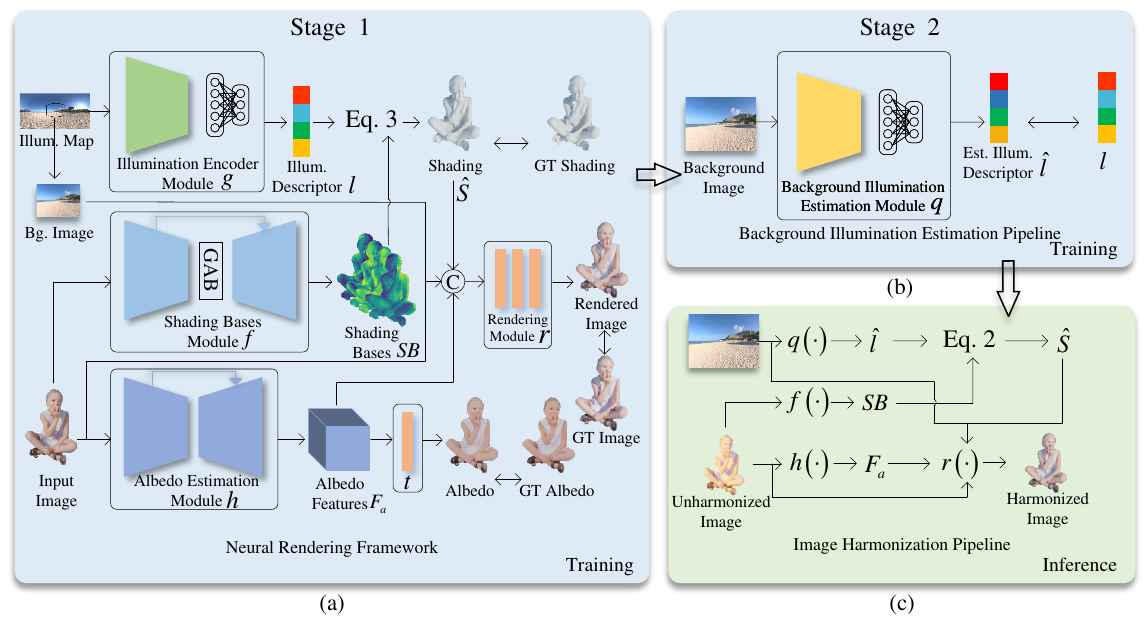}
    \caption{An overview of our proposed image harmonization method. Our method has two training stages: training the Neural Rendering Framework (NRF) and training the Background Illumination Estimation Module (BIEM). The key to the first stage is to learn a shading-aware illumination descriptor, which is then estimated from the background image in the second stage. During inference, our image harmonization pipeline combines partial modules of the NRF $\left\{ f,h,r \right\}$ and the BIEM $q$ to adjust the foreground appearance using the estimated background illumination $\hat{l}$.}
    \label{fig:method}
\end{figure*}

% \subsection{Problem Analysis}
We decompose the image harmonization task into two sub-problems: (1) illumination estimation of background images, and (2) re-rendering of foreground objects. The overall pipeline of the proposed image harmonization algorithm is illustrated in Fig.~\ref{fig:method}. We first train a Neural Rendering Framework to learn the shading-aware illumination descriptor in a self-supervised manner (Sec.~\ref{sec:nid}). Then we train a Background Illumination Estimation Module to estimate the shading-aware illumination descriptor from the background image (Sec.~\ref{sec:bie}). The inference pipeline of image harmonization is briefly introduced in Sec.~\ref{sec:ihp}. Finally, we elaborate on the training and implementation details in Sec.~\ref{sec:train}.

\subsection{Neural Rendering Framework}\label{sec:nid}
As shown in Fig.~\ref{fig:method}(a), the Neural Rendering Framework is composed of three neural network modules and one rendering module. First, the Shading Bases Module and the Illumination Encoder Module generate the shading using the input foreground image and the illumination map. Then, the Albedo Estimation Module makes an estimate of the albedo from the input image. Finally, the Rendering Module combines the albedo feature, the shading, the background image and the input image to re-render the input image under a novel illumination. Below we describe these modules in detail.

\textbf{Shading Bases Module.} Inspired by the illumination cone theory \cite{belhumeur1998set}, the Shading Bases Module $f$, parameterized by $\theta_{f}$, is designed to generate a set of $K$ shading bases $SB \in \mathbb{R}^{K \times H \times W}$, given the input foreground image $\tilde{I} \in \mathbb{R}^{3\times H \times W}$,

\begin{equation}
SB=f(\tilde{I} ;{\theta }_{f}).
\label{eq: shadingmodule}
\end{equation}

The Shading Bases Module, based on a U-Net architecture \cite{ronneberger2015u}, consists of a downsampling sub-module and a upsampling sub-module. The downsampling sub-module is mainly composed of a series of Residual Dense Blocks \cite{zhang2018residual} (RDBs) followed by max-pooling layers. The upsampling sub-module is composed of several convolution layers and upsampling layers. In addition, we utilize the Global Attention Block (GAB) at the bottleneck of the downsampling sub-module to imitate long-range interactions between distant pixels in global illumination. The GAB is composed of 6 transformer layers \cite{vaswani2017attention}.

\textbf{Illumination Encoder Module.} The purpose of the Illumination Encoder Module $g$ is to encode the illumination map $L \in \mathbb{R}^{3 \times {H}' \times {W}'}$ as a low dimensional illumination descriptor $l \in \mathbb{R}^{3 \times K}$,

\begin{equation}
l=g(L;{\theta }_{g}),
\label{eq: illumencodermodule}
\end{equation}
where $K\ll {H}'\times {W}'$. In addition, in order to simultaneously perceive shading and illumination distribution, our illumination descriptor combines different shading bases to generate the final shading $\hat{S} \in \mathbb{R}^{3 \times H \times W}$ which contains global  illumination effects,

\begin{equation}
\hat{S}_{c i j}=\sum_{k=1}^{K} l_{c k} \times S B_{k i j}.
\end{equation}

The network architecture of the Illumination Encoder Module is similar to the downsampling sub-module of the Shading Bases Module. The only difference is that the transformer layers are replaced by three fully-connected layers for outputting the illumination descriptor, which further reduces the amount of network parameters. Note that the first two fully-connected layers are both followed by a rectified linear activation function.

\textbf{Albedo Estimation Module.} The Albedo Estimation Module $h$ is designed to extract the albedo feature $F_{a} \in \mathbb{R}^{C\times H \times W}$ from the input foreground image $\tilde{I}$,
\begin{equation}
F_{a}=h(\tilde{I};{\theta }_{h}).
\end{equation}
Then, one convolution layer $t$ that takes $F_{a}$ as input is adopted to estimate the albedo $\hat{A} \in \mathbb{R}^{3\times H \times W}$: $\hat{A}=t\left( {{F}_{a}};{{\theta }_{t}} \right)$.

The network architecture of the Albedo Estimation Module is the same as that of the Shading Bases Module without GAB. In addition, the channel number of RDBs is reduced by half.

\textbf{Rendering Module.} After obtaining the albedo feature and the shading, the Rendering Module $r$ performs the final rendering,
\begin{equation}
\hat{I}=r(F_{a}, \hat{S}, \tilde{I}, B;{\theta }_{r}),
\end{equation}
where $B$ denotes the background image. In order to preserve details in the foreground, the input image $\tilde{I}$ is also fed to the Rendering Module. Note that the output image $\hat{I}$ shares the same content with the input image $\tilde{I}$ but under a different illumination condition $L$.

The network architecture of the Rendering Module is the same as that of the Albedo Estimation Module.

\subsection{Background Illumination Estimation Module}\label{sec:bie}
Once we have obtained the shading-aware illumination descriptor via the Neural Rendering Framework, the goal of the Background Illumination Estimation Module $q$, which is shown in Fig.~\ref{fig:method} (b), is to estimate the illumination descriptor given the input background image,

\begin{equation}
\hat{l}=q(B ; \theta_{q}).
\label{eq:biem}
\end{equation}

The Background Illumination Estimation Module (BIEM) shares the same network architecture with the Illumination Encoder Module (IEM). However, there are two main differences between BIEM and IEM. First, the inputs of IEM and BIEM are different. As shown in Eq. \ref{eq: illumencodermodule} and Eq. \ref{eq:biem}, the input of IEM is the illumination map $L$, while the input of BIEM is the background image $B$. Note that the illumination map is not available in the inference stage of image harmonization and only the background image is used. Second, the key to the first training stage is to train an IEM to compress a high-dimensional illumination map into a low-dimensional shading-aware illumination descriptor. Once the first training stage is finished, the pre-trained IEM will be later used in the second training stage to supervise the training of the BIEM. In other words, we use the BIEM to estimate an illumination descriptor from a background image, where the ground-truth illumination descriptor is provided by the pre-trained IEM.

Refer to the supplementary materials for more implementation details of all network structures.

\input{Tables/sota}

\begin{figure*}[h]
    \centering
    \includegraphics[width=0.98\linewidth]{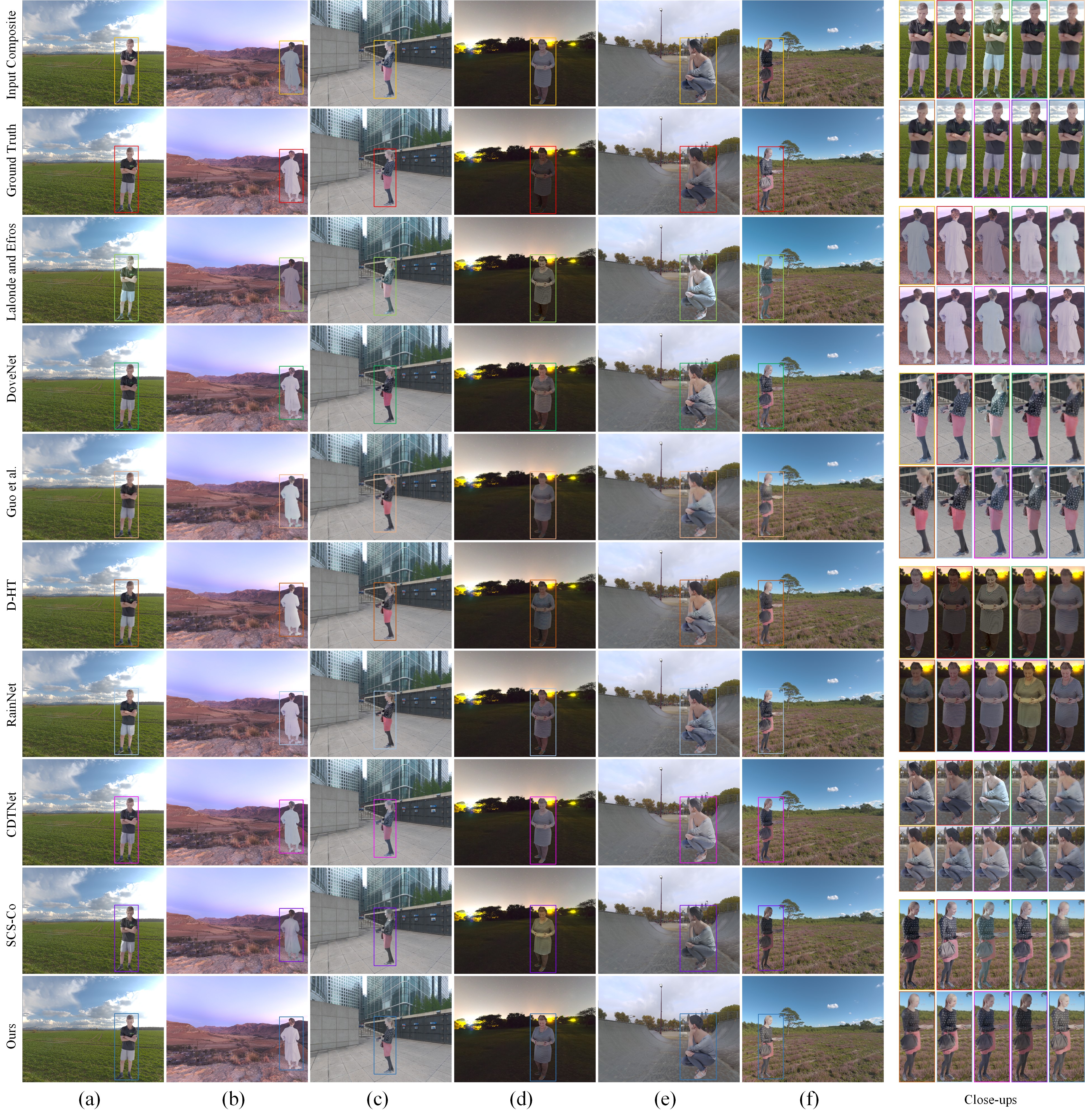}
    \caption{Qualitative comparison of different methods on our test set. We show representative examples with close-up details focusing on shading variations. Our method outperforms all other approaches with more accurate and sharper results.}
    \label{fig:sota}
\end{figure*}

\subsection{Image Harmonization Pipeline}\label{sec:ihp}
As shown in Fig.~\ref{fig:method}(c), our image harmonization pipeline consists of two modules, namely Background Illumination Estimation Module $q$ and Foreground Rendering Module $\left\{ f,h,r \right\}$.

The Foreground Rendering Module leverages partial modules of the existing Neural Rendering Framework to re-render the input unharmonized foreground image to make it more compatible with the background image.

\subsection{Training Details}\label{sec:train}
We train the model on our image harmonization dataset with ground truth $\{I, A, S, L\}$, where $I, A, S$ denote the harmonized image, albedo and shading respectively. Our training is divided into two stages: training the Neural Rendering Framework and training the Background Illumination Estimation Module.

At the first stage, we train the Neural Rendering Framework. The $\mathcal{L}_{1}$ loss is applied for shading, albedo and the output image. In addition, inspired by \cite{zhao2016loss}, the $\mbox{SSIM}$ metric is utilized to encourage the neural network to produce visually pleasing images. Thus, the loss ${{L}_{NR}}$ for the Neural Rendering Framework is defined as,
\begin{equation}\small
\begin{split}
L_{NR}=\|S-\hat{S}\|_{1}+\|A-\hat{A}\|_{1}+\|I-\hat{I}\|_{1}+\lambda(1-\operatorname{SSIM}(S, \hat{S}))\\
+\lambda(1-\operatorname{SSIM}(A, \hat{A}))+\lambda(1-\operatorname{SSIM}(I, \hat{I})),
\end{split}
\end{equation}
where the weight ${\lambda}$ is set to $1$ in our experiments.

At the second stage, we train the Background Illumination Estimation Module. The $\mathcal{L}_{1}$ loss is used for the illumination descriptor. Also, the predicted illumination descriptor and the shading bases are utilized to render the shading and then minimize the error between the rendered shading and the ground truth shading. The loss $L_{B I E}$ for the Background Illumination Estimation Module is defined as,
\begin{equation}
L_{B I E}=\|l-\hat{l}\|_{1}+\left\|S-\sum_{k} \hat{l}_{c k} \times S B_{k i j}\right\|_{1}.
\end{equation}

\section{Experiments}
To validate the effectiveness of our image harmonization pipeline, we first compare our method with several state-of-the-art methods. Then we compare our neural illumination descriptor against the common illumination representation (i.e., HDR illumination maps) and our neural shading bases against the spherical harmonic bases to demonstrate their advantages in terms of rendering quality. A user study on real data is also conducted to confirm the effectiveness of our method. Finally, we perform extensive ablation studies to illustrate the contribution of each component of our framework in isolation.

\subsection{Experimental Setup}

\textbf{Evaluation metrics.} We evaluate the realism of harmonized images using fMAE, fPSNR, fSSIM \cite{wang2004image} and LPIPS \cite{zhang2018perceptual}, where the prefix f indicates that the metric measurement is calculated only using the foreground region.

\textbf{Baselines.} We compare with one traditional method \cite{lalonde2007using} and six deep learning-based methods \cite{Cong2020DoveNet,guo2021iih,ling2021region,guo2021dht,cong2022high,hang2022scs}. For deep learning-based methods, we select recent open-source methods \cite{Cong2020DoveNet,guo2021iih,ling2021region,guo2021dht} achieving state-of-the-art performance. In addition, Cong et al. \cite{cong2022high} provided us with their code and pre-trained model. For a fair comparison, we re-train their models on our image harmonization dataset according to the experiment settings given by the authors. We report their results when the training losses converge. Refer to the supplementary materials for more experimental details. The results of SCS-Co~\cite{hang2022scs} are provided by the authors.

\subsection{Comparison with State-of-the-art}

\textbf{Quantitative results.} Tab.~\ref{tab:sota} summarizes the quantitative results obtained by our method as well as the competing methods. Our method achieves the best results on the \textit{sunny} and \textit{cloudy} scenes, which can be attributed to its ability to generate realistic shadings. However, our method gets lower scores on the \textit{night} scene compared to previous works. This is primarily due to the fact that the night images lack noticeable shading variations. Overall, our method achieves the best performance in all metrics when using the entire test set for evaluation. 
In addition, compared with other learning-based methods, CDTNet specially integrates with the color mapping module. However, this module cannot handle shading variations and may result in limited performance. Since SCS-Co does not consider the perception of illumination and its training data only contains variations in brightness and color, its performance is severely degraded on our test data which also contains shading variations.

We demonstrate the effect of using the illumination maps as inputs to extract the illumination descriptors. As can be observed from Tab.~\ref{tab:sota}, using the illumination maps as inputs (ours w/ illum. maps) leads to a significant increase in the rendering performance.

We also compare our method against the baselines using the number of parameters. Despite its complexity, our entire framework has a total of 10.403M parameters, which is approximately one-fifth of the amount of the second-best baseline with 50.763M parameters.

\textbf{Qualitative results.}
Harmonized images produced by different methods are compared in Fig.~\ref{fig:sota}. We display the qualitative results with different lighting conditions on several scenes, including \textit{sunny}, \textit{cloudy}, \textit{sunrise/sunset}, and \textit{night}.
Our method produces compelling results that are closer to the ground truth in terms of photo-realism. For instance, in the first column of Fig.~\ref{fig:sota}, there is an observable illumination inconsistency between the foreground and the background in the input composite image.
Specifically, the background suggests that the main illumination source is located at the rear right, whereas, the foreground appears to be illuminated from the left.
The result of Lalonde et al.~\cite{lalonde2007using} shows greenish colors, and all the other comparative methods \cite{guo2021iih,ling2021region,Cong2020DoveNet,guo2021dht,cong2022high,hang2022scs} basically retain the original illumination (e.g., the boy neck in close-ups). In contrast, our method consistently relights the foreground object, making it more consistent with the background illumination.

In the fifth column of Fig.~\ref{fig:sota}, the foreground object in the input image appears to be illuminated from the right, whereas the background is a cloudy image.
Ideally, under such background illumination, the foreground object should appear smooth lighting. The result of Lalonde and Efros~\cite{lalonde2007using} is inconsistent in terms of both color and illumination. The results of CDTNet~\cite{cong2022high} and SCS-Co~\cite{hang2022scs} almost completely preserve the effect of the original lighting. Although RainNet~\cite{ling2021region}, DoveNet~\cite{Cong2020DoveNet}, Guo et al.~\cite{guo2021iih} and D-HT\cite{guo2021dht} produce the results that are a step closer to the ground truth, the highlights on the woman's left arm are improperly preserved.
Our method not only effectively delights the foreground object, but also re-renders it under a smooth illumination.

\begin{figure}[h]
    \centering
    \includegraphics[width=0.98\linewidth]{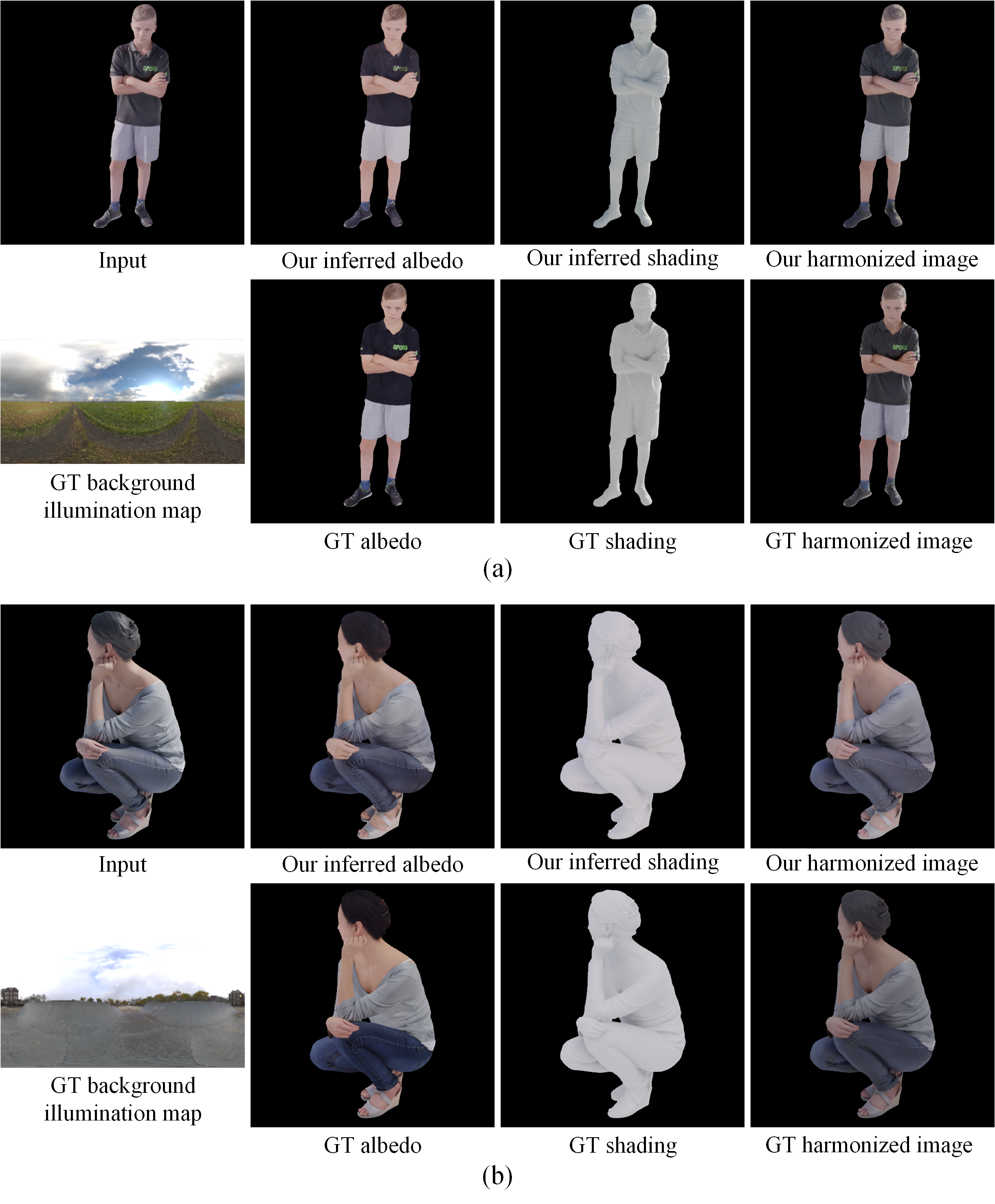}
    \caption{Visualization of our intermediate results. Note that our inferred albedos effectively remove the original illumination effects, and our inferred shadings correctly contain the effects of the background illuminations.}
    \label{fig:intermedia_results}
\end{figure}

\textbf{Effects of the inferred shading and albedo.}
For the single image harmonization task, there are two challenges: (1) removing the original illumination on the foreground and (2) generating the shadings under the background illumination. In this paper, we design the Albedo Estimation Module and the Shading Bases Module to solve these two problems respectively. As shown in Fig.~\ref{fig:intermedia_results}, our inferred albedos effectively remove the original illumination effects, and our inferred shadings correctly contain the effects of the background illuminations. As a result, our harmonized images are more realistic and physically correct. In contrast, those image-to-image harmonization methods perform poorly on these two aspects. As shown in Fig.~\ref{fig:sota}(a)(e), for example, the original light on the boy's left nose and on the woman's clothes are not well eliminated, and even artifacts are introduced in these areas. Moreover, since neither explicit shading modeling nor light perception is conducted, these methods fail to generate plausible shadings.

\subsection{Comparison with HDR Illumination Map}\label{sec:comp_with_hdr}
\input{Tables/Illum_Resp}
The efficacy of our learned illumination descriptor is compared with the HDR illumination map for light estimation. Specifically, we train an encoder-decoder based neural network to map the background image to its corresponding panoramic HDR illumination map. We train this network until its loss converges. Then, we use it to estimate the HDR panoramic image from the background image. The estimated HDR image is further used as part of our Neural Rendering Framework. The rendered images are compared against those generated using our learned illumination descriptor, which is reported in Tab.~\ref{tab:ablations}. Note that our learned illumination descriptors achieve obviously better performance compared to the estimated HDR images. In fact, it is very difficult to accurately estimate the HDR illumination map from the background image due to its huge amount of parameters.

\subsection{Comparison with Spherical Harmonic Bases}
\begin{figure}[htb]
    \centering
    \includegraphics[width=.99\linewidth]{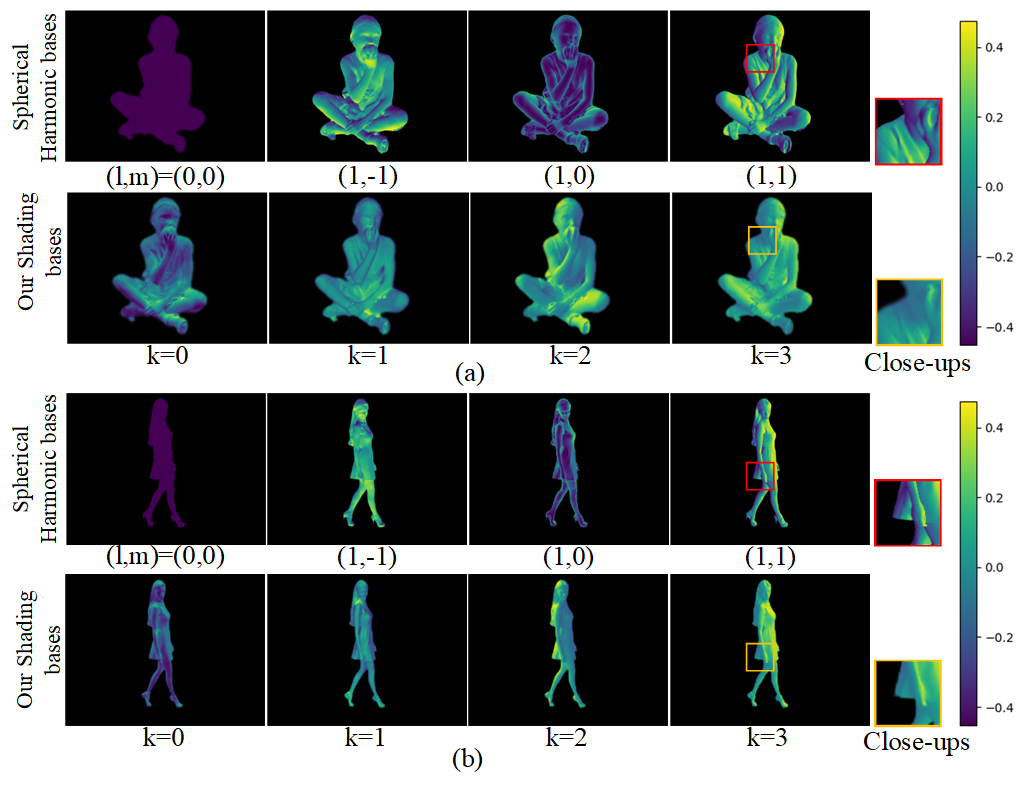}
    \caption{Comparison with Spherical Harmonic bases.}
    \label{fig:shading_bases}
\end{figure}
In Fig.~\ref{fig:shading_bases}, we compare our shading bases against the Spherical Harmonics (SH) bases~\cite{ramamoorthi2001efficient}. Our main objective is to emphasize the advantage of our shading bases in comparison to the SH bases in terms of generating cast shadows.

We generate the first 4 SH bases ${{Y}_{lm}}$ (with (l,m) = \{(0,0), (1,-1), (1,0), (1,1)\}, where $l \geq 0$ and $-l \leq m \leq l$) for comparison. $(0,0)$ indicates ambient illumination and has no specific illumination direction. $(1,-1)$, $(1,0)$ and $(1,1)$ show that the light source is located below, behind and to the left of the little girl, respectively.

For this experiment, we set $K$ of the Shading Bases Module to 4. We visualize the shading bases learned by our Shading Bases Module in Fig.~\ref{fig:shading_bases}, where $k$ is used to denote the $k_{th}$ shading basis. The values $k=0, 1, 2, 3$ indicate that the light source is located behind, in front, to the right, and to the left of the little girl, respectively.

In comparison to the SH bases, our shading bases contain the cast shadow effects which are explicitly omitted by the spherical harmonic bases. Taking the last column of Fig.~\ref{fig:shading_bases} as an example, both our shading basis and the spherical harmonics basis are illuminated from the right. From the close-ups in Fig.~\ref{fig:shading_bases}(a), it can be observed that the shoulder region is occluded by the head. Spherical harmonics however produce bright intensities without cast shadows. While our shading bases contain cast shadows that are congruent with the illumination direction, as shown in the close-ups.

\begin{figure*}[htb]
    \centering
    \includegraphics[width=0.75\linewidth]{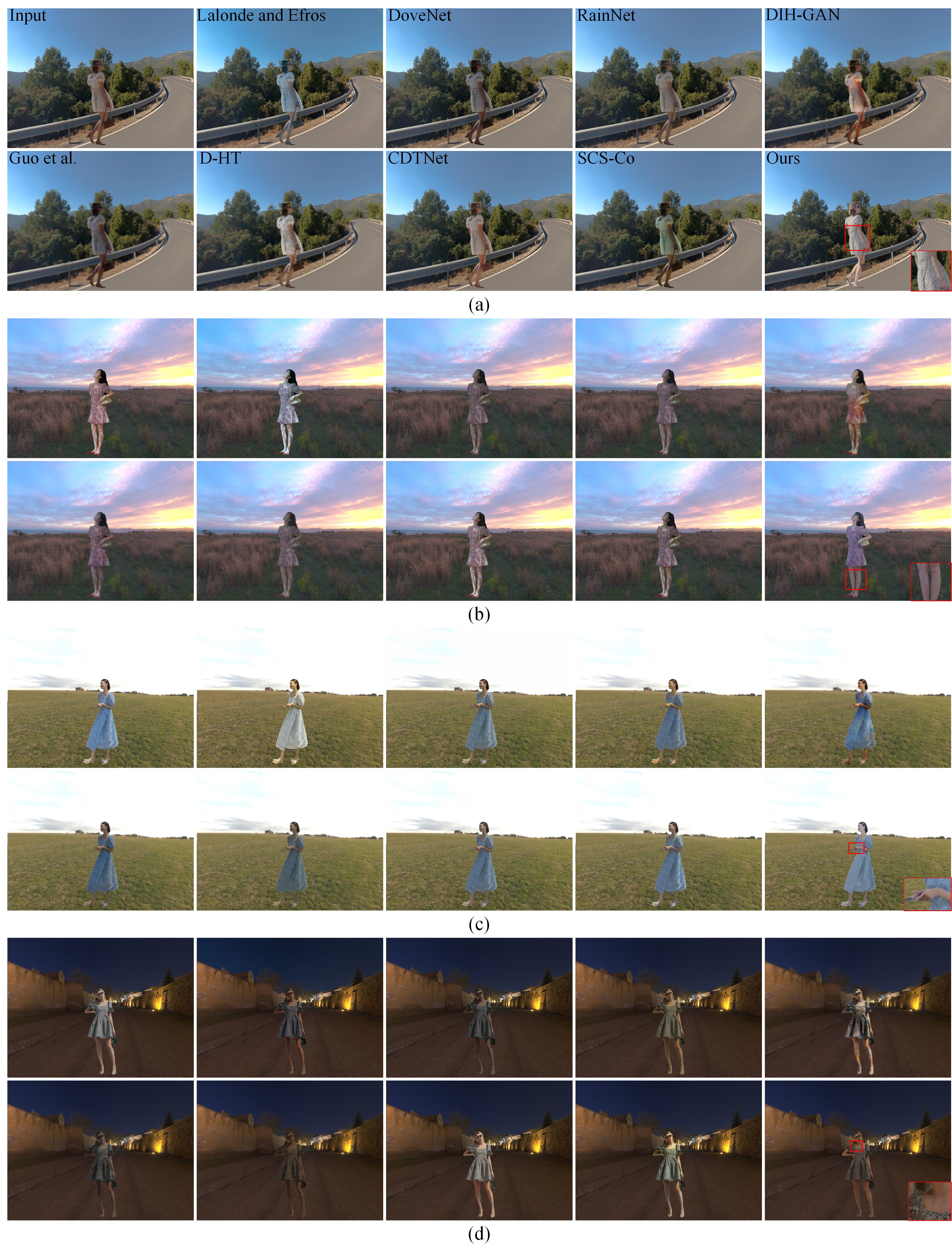}
\vspace{-2mm}
    \caption{Qualitative comparison of different methods on real data across different weather conditions. The details in red boxes show that our method is capable of generating plausible shadings which are consistent with the background illumination, while other methods can only adjust the color and brightness of the foreground.}
    \label{fig:user-study}
\end{figure*}

\subsection{User Study on Real Data}
We also conduct user study on real data to validate the performance of our proposed method. We made 58 composite images of which both the foreground images and the background images are collected from the Internet. Specifically, foreground humans are collected from Taobao\cite{taobao} and captured by real cameras for clothes display. The background images are collected from Poly Haven \cite{polyHaven} and HDR MAPS \cite{hdrmaps}, which are all captured by professional digital cameras. We will make this benchmark dataset publicly available. For deep learning-based methods, we compare against the state-of-the-art methods, namely DoveNet \cite{Cong2020DoveNet}, RainNet \cite{ling2021region}, DIH-GAN \cite{bao2022deep}, CDTNet \cite{cong2022high}, SCS-Co \cite{hang2022scs}, Guo et al. \cite{guo2021iih} and D-HT \cite{guo2021dht}. Note that for DoveNet, RainNet, CDTNet, Guo et al. and D-HT, we use their released pre-trained models to process the input composite images. The results of DIH-GAN and SCS-Co are provided by the authors. We also compare against the traditional method proposed by Lalonde and Efros \cite{lalonde2007using}. For each composite image processed by these nine methods, we ask 27 individuals to score the visual quality. As inspired by \cite{jiang2021ssh}, the following three questions are considered for scoring: (1) Are the brightness and color of the foreground and background consistent; (2) Are the illumination directions of the foreground and background consistent; and (3) Are the texture distortions/artifacts of the foreground serious. The visual quality score ranges from 0 to 3 (worst to best quality). Tab. \ref{tab:user-study} reports the results. It shows that we achieve a large advantage on question 2, which is mainly due to the fact that neither previous methods \cite{Cong2020DoveNet, ling2021region,guo2021iih,guo2021dht,cong2022high} nor their corresponding training data have yet considered shading variations. Besides, the recently proposed DIH-GAN \cite{bao2022deep} does not explicitly model foreground shading, and many of their 3D models are created by CG software, which results in poor generalization to real data. As shown in Fig. \ref{fig:user-study}, 
the details in red boxes show that our method is able to generate foreground shadings that are consistent with the background illumination. In contrast, other methods fail to generate plausible shading and even erroneously transfer the color of the background objects (e.g., the grasses) to the foreground due to the lack of perception of the illumination in the background.

\input{Tables/user_study}

\subsection{Generalization to Indoor Scenes and Non-Human Objects}
\begin{figure}[ht]
    \centering
    \includegraphics[width=\linewidth]{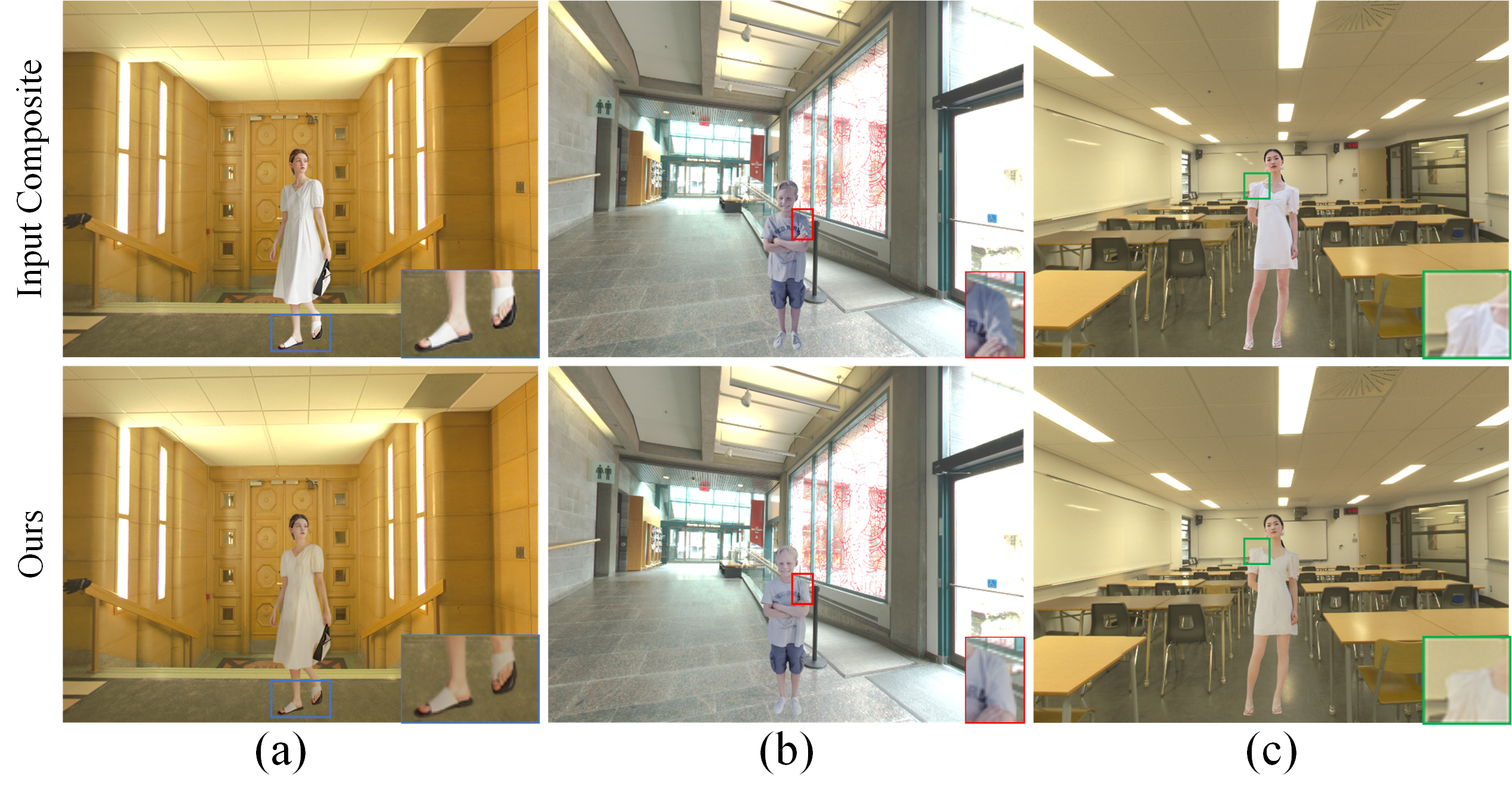}
      \caption{Generalization to indoor scenes. Zoom in for more details.}
    \label{fig:indoor-scenes}
\end{figure}

Fig.~\ref{fig:indoor-scenes} shows the generalization of our method to indoor scenes. First, our method is able to perceive illumination for indoor scenes, especially the illumination direction. For example, in Fig.~\ref{fig:indoor-scenes} (b), the background image indicates that the primary light source in the scene comes from the right side (i.e., the windows), and our generated shading (e.g., the details in the red box) is consistent with the direction of the primary light source in the scene. Second, as shown in Fig.~\ref{fig:indoor-scenes} (a), our result, especially the details in the blue box, is more realistic owing to the appropriate illumination brightness and color. This can also be observed in Fig.~\ref{fig:indoor-scenes} (c).

However, our method does not yet account for spatially varying illumination estimation which can further improve the realism for indoor scene harmonization. In the future, one of the potential solutions is to estimate an illumination descriptor for individual background image pixel.

\begin{figure}[ht]
    \centering
    \includegraphics[width=0.97\linewidth]{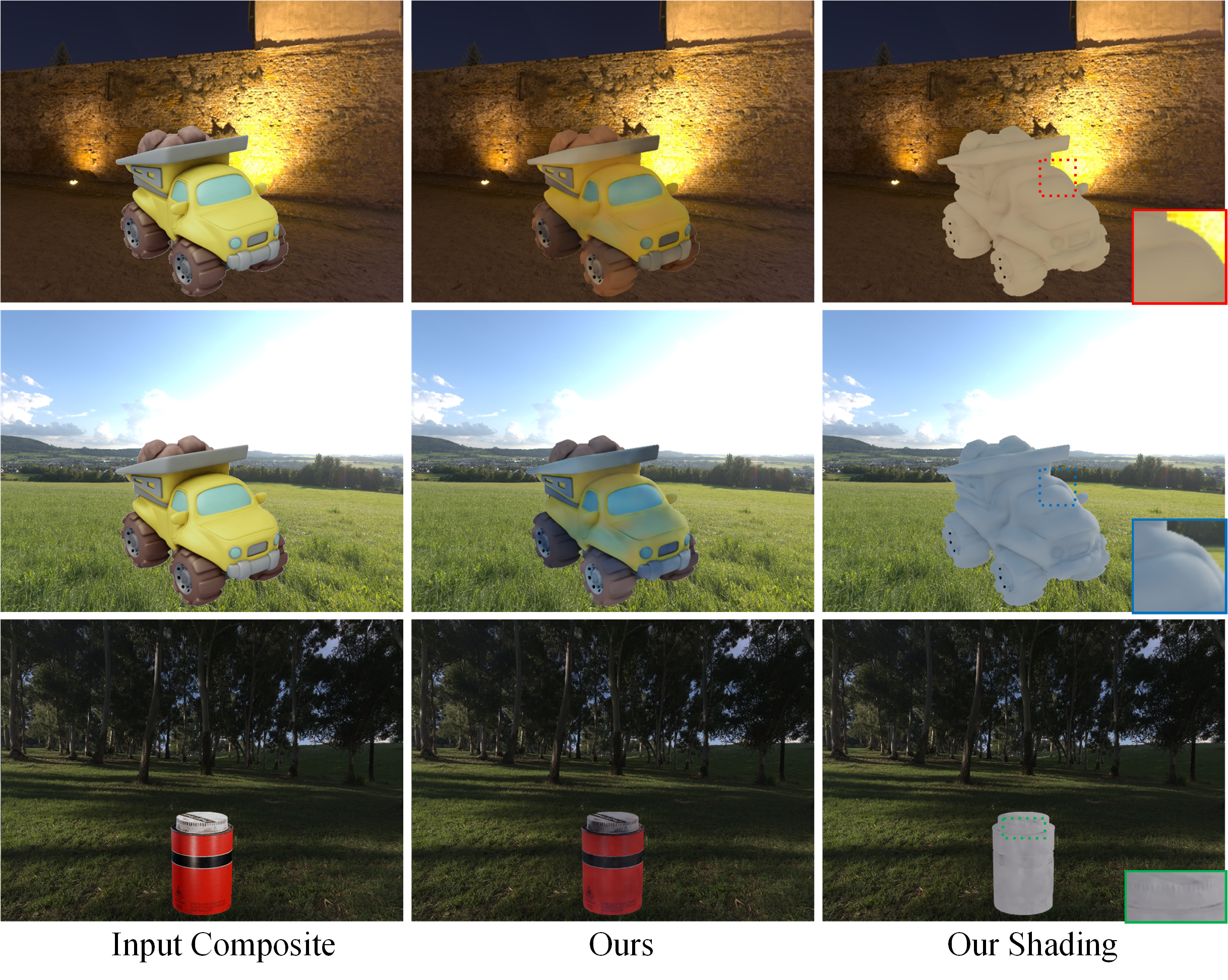}
      \caption{Generalization to non-human objects.}
    \label{fig:non-humans}
\end{figure}

Fig.~\ref{fig:non-humans} shows the generalization of our method to non-human objects. Although the constructed training set only covers human objects, our approach generalizes well to non-human objects, specifically generating reasonable shadings that are consistent with the target background lighting. For example, in the second row of Fig.~\ref{fig:non-humans}, it can be observed from the background image that the sun is located behind the right side of the toy car. Not only is our harmonized toy car mostly backlit, but its right side is partially illuminated by the sun, as shown in the blue box. In addition, adding more different types of objects to the training set could further improve the generalization performance of the proposed method.

\subsection{Ablation Study}
The ablation study is conducted to demonstrate the effectiveness of each component on the Neural Rendering Framework (NRF).
\input{Tables/AblationNRF}

\textbf{Neural rendering framework Ablation.}
We demonstrate how the use of the albedo features $F_a$, the input unharmonized image $\hat{I}$ and the background image $B$ as additional inputs to the Rendering Module can improve the overall performance. We also perform an ablation on the loss function of our Neural Rendering Framework. Quantitative results are reported in Tab.~\ref{tab:ablation_NRF}.

We start with the baseline NRF which uses the concatenated shading and albedo as inputs to the Rendering Module. Replacing the albedo image with the albedo feature $F_a$ results in better performance. We attribute this improvement to the albedo feature $F_a$ which contains richer information.

We then proceed to add the input unharmonized image $\hat{I}$ in conjunction with the albedo feature $F_a$ and the shading as inputs to the Rendering Module, and observe a slight improvement in performance.
The delighting, which occurs as a consequence of albedo estimation, can result in a loss of information in the final rendered image. Therefore, using the unharmonized image $\hat{I}$ as additional input can make up for the lost information.
When the background image is also added to the Rendering Module, there is a moderate increase in performance. In fact, by exploiting the brightness and color information of the background image, the Rendering Module is able to generate the foreground appearance more accurately.

Finally, in addition to the $\mathcal{L}_{1}$ loss of the baseline NRF, the SSIM loss function is added. Experimental results show that adding the SSIM loss significantly improves the performance of our framework, especially in terms of the foreground SSIM (fSSIM) and LPIPS metric measurements.

\input{Tables/AblationK}
\begin{figure}[ht]
    \centering
    \includegraphics[width=\linewidth]{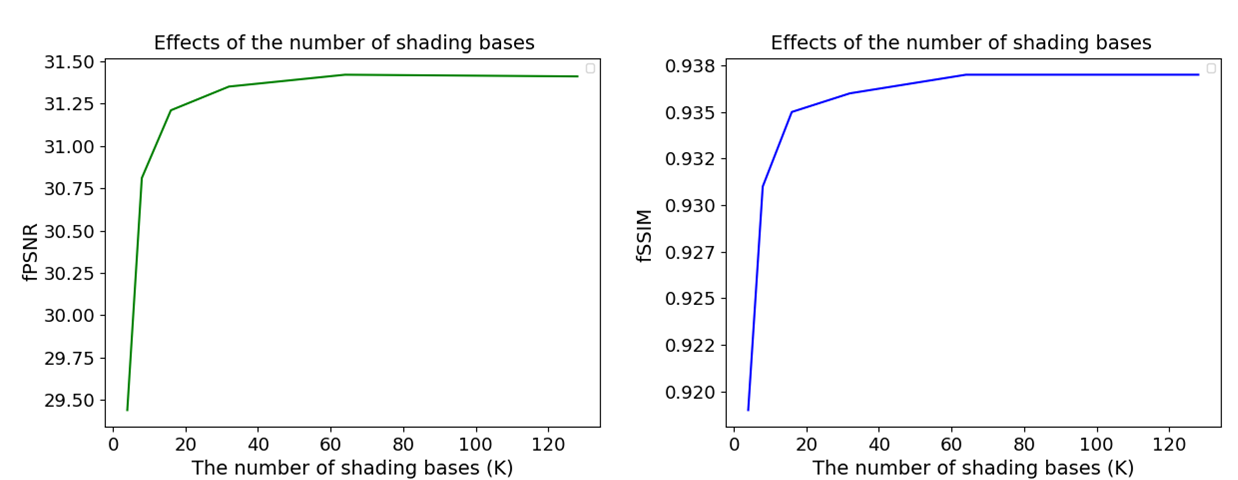}
      \caption{Effects of the number of shading bases.}
    \label{fig:ablation_K}
\end{figure}

\textbf{Effects of the number of shading bases ($K$).}

To measure the influence of the number of shading bases on the performance of our Shading Bases Module, we train the Shading Bases Module with 6 different values of $K$ (4, 8, 16, 32, 64, 128). The results are reported in Tab.~\ref{tab:ablation_k}. We also visualize them in Fig.~\ref{fig:ablation_K}. Note that the reported metrics (fPSNR, fSSIM) are obtained by comparing the results generated by our Shading Bases Module against the corresponding ground truth shading images. There exists a significant increase in the performance when the value of $K$ increases from 4 to 32, then the increase becomes less obvious when $K>32$. To balance between the rendering performance and the computational complexity, we resolve to use $K=32$ as the optimal number of shading bases.

\subsection{Discussions}
\textbf{Impact of object placement.} Object placement aims to place the foreground within the background image with a suitable location and size. From the view of physical image formation, the most important factor affecting image harmonization is the location of objects. Because different locations may have different lighting, depending on the type of scene. Especially in indoor scenes, the lighting at different locations may vary greatly. Therefore, the acquisition of illumination at different locations in scenes and its accurate estimation pose a greater challenge to image harmonization. In the near future, we will continue to focus on image harmonization with spatially-varying lighting estimation.

\section{Conclusions}

In this paper, we have contributed a large-scale photo-realistic image harmonization dataset involving variations in color, brightness,
and shading. In addition, a novel Neural Rendering Framework is designed to learn a shading-aware illumination descriptor from the illumination maps. A neural Shading Bases Module is proposed to generate the foreground shading using the shading-aware illumination descriptor estimated from the background. Extensive experiments on the self-constructed dataset and real data demonstrate the effectiveness of our proposed method.

\textbf{Limitations.} This work has several limitations that can be further improved. At now, we focus on one specific object type (i.e. human body), which limits the application scope of our method. Extending to different types could improve the ability to generalize across a wide spectrum of objects. Additionally, only Lambertian objects are considered. When introducing the specular BRDFs \cite{ashikmin2000microfacet}, our model could be applied to the objects with specular reflection.

% \section*{Acknowledgments}
% This should be a simple paragraph before the References to thank those individuals and institutions who have supported your work on this article.

% {\appendix[Proof of the Zonklar Equations]
% Use $\backslash${\tt{appendix}} if you have a single appendix:
% Do not use $\backslash${\tt{section}} anymore after $\backslash${\tt{appendix}}, only $\backslash${\tt{section*}}.
% If you have multiple appendixes use $\backslash${\tt{appendices}} then use $\backslash${\tt{section}} to start each appendix.
% You must declare a $\backslash${\tt{section}} before using any $\backslash${\tt{subsection}} or using $\backslash${\tt{label}} ($\backslash${\tt{appendices}} by itself
%  starts a section numbered zero.)}


\bibliographystyle{IEEEtran}
\bibliography{IEEEabrv, ImageHarmonization}

\vfill

\end{document}

%% file: Tables/dataset_statistics.tex
\begin{table}[ht]
% \resizebox{\linewidth}{!}{
\caption{The number of training and test images on each scene.}
\begin{tabular}{c|cccc|c}
\hline
Scene & Sunny & Sunrise/Sunset & Cloudy & Night & All    \\ \hline
\#Train       & 62,074 & 27,471          & 48,051  & 5,794 & 143,390 \\
\#Test        & 10,472 & 3,608           & 7,508   & 460   & 22,048  \\ \hline
\end{tabular}
\label{tab:dataset statistics}
\end{table}

%% file: Tables/sota.tex
% \usepackage{multirow}
% \usepackage[normalem]{ulem}
\begin{table*}[]
\caption{Quantitative evaluation of different methods on the test set. The best results are marked in bold. The second best results are underlined.  Our method achieves the best results on the entire test set with the fewest parameters.}
\centering
\resizebox{0.99\linewidth}{!}{
\begin{tabular}{|
>{\columncolor[HTML]{FFFFFF}}c |
>{\columncolor[HTML]{FFFFFF}}c |
>{\columncolor[HTML]{FFFFFF}}c |
>{\columncolor[HTML]{FFFFFF}}c |
>{\columncolor[HTML]{FFFFFF}}c |
>{\columncolor[HTML]{FFFFFF}}c |
>{\columncolor[HTML]{FFFFFF}}c |
>{\columncolor[HTML]{FFFFFF}}c |
>{\columncolor[HTML]{FFFFFF}}c |
>{\columncolor[HTML]{FFFFFF}}c |
>{\columncolor[HTML]{FFFFFF}}c 
>{\columncolor[HTML]{FFFFFF}}c |}
\hline
{\color[HTML]{000000} Sub-dataset} &
  {\color[HTML]{000000} \begin{tabular}[c]{@{}c@{}}Evaluation \\ Metric\end{tabular}} &
  {\color[HTML]{000000} \begin{tabular}[c]{@{}c@{}}Input \\ Composite\end{tabular}} &
  {\color[HTML]{000000} \begin{tabular}[c]{@{}c@{}}Lalonde and\\ Efros \cite{lalonde2007using}\end{tabular}} &
  {\color[HTML]{000000} \begin{tabular}[c]{@{}c@{}}DoveNet\\ \cite{Cong2020DoveNet}\end{tabular}} &
  {\color[HTML]{000000} \begin{tabular}[c]{@{}c@{}}Guo et al.\\ \cite{guo2021iih}\end{tabular}} &
  {\color[HTML]{000000} \begin{tabular}[c]{@{}c@{}}RainNet\\ \cite{ling2021region}\end{tabular}} &
  {\color[HTML]{000000} \begin{tabular}[c]{@{}c@{}}D-HT\\ \cite{guo2021dht}\end{tabular}} &
  {\color[HTML]{000000} \begin{tabular}[c]{@{}c@{}}CDTNet\\ \cite{cong2022high}\end{tabular}} &
  {\color[HTML]{000000} \begin{tabular}[c]{@{}c@{}}SCS-Co\\ \cite{hang2022scs}\end{tabular}} &
  \multicolumn{1}{c|}{\cellcolor[HTML]{FFFFFF}{\color[HTML]{000000} Ours}} &
  {\color[HTML]{000000} \begin{tabular}[c]{@{}c@{}}Ours w/\\  Illum. map\end{tabular}} \\ \hline
\cellcolor[HTML]{FFFFFF}{\color[HTML]{000000} } &
  {\color[HTML]{000000} fMAE$\downarrow$} &
  {\color[HTML]{000000} 0.099} &
  {\color[HTML]{000000} 0.110} &
  {\color[HTML]{000000} 0.073} &
  {\color[HTML]{000000} 0.076} &
  {\color[HTML]{000000} 0.067} &
  {\color[HTML]{000000} 0.074} &
  {\color[HTML]{000000} 0.091} &
  {\color[HTML]{000000} 0.119} &
  \multicolumn{1}{c|}{\cellcolor[HTML]{FFFFFF}{\color[HTML]{000000} {\ul 0.062}}} &
  {\color[HTML]{000000} \textbf{0.055}} \\
\cellcolor[HTML]{FFFFFF}{\color[HTML]{000000} } &
  {\color[HTML]{000000} fPSNR$\uparrow$} &
  {\color[HTML]{000000} 19.52} &
  {\color[HTML]{000000} 18.33} &
  {\color[HTML]{000000} 21.38} &
  {\color[HTML]{000000} 21.15} &
  {\color[HTML]{000000} 22.02} &
  {\color[HTML]{000000} 21.47} &
  {\color[HTML]{000000} 20.21} &
  {\color[HTML]{000000} 17.90} &
  \multicolumn{1}{c|}{\cellcolor[HTML]{FFFFFF}{\color[HTML]{000000} {\ul 22.61}}} &
  {\color[HTML]{000000} \textbf{23.77}} \\
\cellcolor[HTML]{FFFFFF}{\color[HTML]{000000} } &
  {\color[HTML]{000000} fSSIM$\uparrow$} &
  {\color[HTML]{000000} 0.804} &
  {\color[HTML]{000000} 0.787} &
  {\color[HTML]{000000} 0.821} &
  {\color[HTML]{000000} 0.851} &
  {\color[HTML]{000000} 0.831} &
  {\color[HTML]{000000} 0.810} &
  {\color[HTML]{000000} 0.811} &
  {\color[HTML]{000000} 0.757} &
  \multicolumn{1}{c|}{\cellcolor[HTML]{FFFFFF}{\color[HTML]{000000} {\ul 0.869}}} &
  {\color[HTML]{000000} \textbf{0.893}} \\
\multirow{-4}{*}{\cellcolor[HTML]{FFFFFF}{\color[HTML]{000000} Sunny}} &
  {\color[HTML]{000000} LPIPS$\downarrow$ ($\times$e-2)} &
  {\color[HTML]{000000} 1.089} &
  {\color[HTML]{000000} 2.353} &
  {\color[HTML]{000000} 1.111} &
  {\color[HTML]{000000} 1.304} &
  {\color[HTML]{000000} 0.815} &
  {\color[HTML]{000000} 1.114} &
  {\color[HTML]{000000} 1.096} &
  {\color[HTML]{000000} 3.182} &
  \multicolumn{1}{c|}{\cellcolor[HTML]{FFFFFF}{\color[HTML]{000000} {\ul 0.791}}} &
  {\color[HTML]{000000} \textbf{0.678}} \\ \hline
\cellcolor[HTML]{FFFFFF}{\color[HTML]{000000} } &
  {\color[HTML]{000000} fMAE$\downarrow$} &
  {\color[HTML]{000000} 0.083} &
  {\color[HTML]{000000} 0.119} &
  {\color[HTML]{000000} 0.071} &
  {\color[HTML]{000000} 0.072} &
  {\color[HTML]{000000} {\ul 0.060}} &
  {\color[HTML]{000000} 0.070} &
  {\color[HTML]{000000} 0.080} &
  {\color[HTML]{000000} 0.091} &
  \multicolumn{1}{c|}{\cellcolor[HTML]{FFFFFF}{\color[HTML]{000000} 0.061}} &
  {\color[HTML]{000000} \textbf{0.058}} \\
\cellcolor[HTML]{FFFFFF}{\color[HTML]{000000} } &
  {\color[HTML]{000000} fPSNR$\uparrow$} &
  {\color[HTML]{000000} 21.48} &
  {\color[HTML]{000000} 18.07} &
  {\color[HTML]{000000} 22.35} &
  {\color[HTML]{000000} 22.49} &
  {\color[HTML]{000000} {\ul 23.60}} &
  {\color[HTML]{000000} 22.30} &
  {\color[HTML]{000000} 21.63} &
  {\color[HTML]{000000} 20.34} &
  \multicolumn{1}{c|}{\cellcolor[HTML]{FFFFFF}{\color[HTML]{000000} 23.38}} &
  {\color[HTML]{000000} \textbf{23.92}} \\
\cellcolor[HTML]{FFFFFF}{\color[HTML]{000000} } &
  {\color[HTML]{000000} fSSIM$\uparrow$} &
  {\color[HTML]{000000} 0.873} &
  {\color[HTML]{000000} 0.820} &
  {\color[HTML]{000000} 0.888} &
  {\color[HTML]{000000} 0.892} &
  {\color[HTML]{000000} 0.899} &
  {\color[HTML]{000000} 0.866} &
  {\color[HTML]{000000} 0.882} &
  {\color[HTML]{000000} 0.843} &
  \multicolumn{1}{c|}{\cellcolor[HTML]{FFFFFF}{\color[HTML]{000000} {\ul 0.926}}} &
  {\color[HTML]{000000} \textbf{0.941}} \\
\multirow{-4}{*}{\cellcolor[HTML]{FFFFFF}{\color[HTML]{000000} Sunrise/Sunset}} &
  {\color[HTML]{000000} LPIPS$\downarrow$ ($\times$e-2)} &
  {\color[HTML]{000000} 0.993} &
  {\color[HTML]{000000} 2.249} &
  {\color[HTML]{000000} 0.915} &
  {\color[HTML]{000000} 1.207} &
  {\color[HTML]{000000} 0.685} &
  {\color[HTML]{000000} 1.008} &
  {\color[HTML]{000000} 1.030} &
  {\color[HTML]{000000} 4.316} &
  \multicolumn{1}{c|}{\cellcolor[HTML]{FFFFFF}{\color[HTML]{000000} {\ul 0.634}}} &
  {\color[HTML]{000000} \textbf{0.517}} \\ \hline
\cellcolor[HTML]{FFFFFF}{\color[HTML]{000000} } &
  {\color[HTML]{000000} fMAE$\downarrow$} &
  {\color[HTML]{000000} 0.084} &
  {\color[HTML]{000000} 0.101} &
  {\color[HTML]{000000} 0.070} &
  {\color[HTML]{000000} 0.074} &
  {\color[HTML]{000000} 0.063} &
  {\color[HTML]{000000} 0.071} &
  {\color[HTML]{000000} 0.082} &
  {\color[HTML]{000000} 0.089} &
  \multicolumn{1}{c|}{\cellcolor[HTML]{FFFFFF}{\color[HTML]{000000} {\ul 0.057}}} &
  {\color[HTML]{000000} \textbf{0.056}} \\
\cellcolor[HTML]{FFFFFF}{\color[HTML]{000000} } &
  {\color[HTML]{000000} fPSNR$\uparrow$} &
  {\color[HTML]{000000} 21.75} &
  {\color[HTML]{000000} 19.38} &
  {\color[HTML]{000000} 22.64} &
  {\color[HTML]{000000} 22.24} &
  {\color[HTML]{000000} 23.67} &
  {\color[HTML]{000000} 22.60} &
  {\color[HTML]{000000} 21.81} &
  {\color[HTML]{000000} 20.64} &
  \multicolumn{1}{c|}{\cellcolor[HTML]{FFFFFF}{\color[HTML]{000000} {\ul 24.14}}} &
  {\color[HTML]{000000} \textbf{24.21}} \\
\cellcolor[HTML]{FFFFFF}{\color[HTML]{000000} } &
  {\color[HTML]{000000} fSSIM$\uparrow$} &
  {\color[HTML]{000000} 0.881} &
  {\color[HTML]{000000} 0.843} &
  {\color[HTML]{000000} 0.899} &
  {\color[HTML]{000000} 0.905} &
  {\color[HTML]{000000} 0.908} &
  {\color[HTML]{000000} 0.873} &
  {\color[HTML]{000000} 0.893} &
  {\color[HTML]{000000} 0.852} &
  \multicolumn{1}{c|}{\cellcolor[HTML]{FFFFFF}{\color[HTML]{000000} {\ul 0.935}}} &
  {\color[HTML]{000000} \textbf{0.947}} \\
\multirow{-4}{*}{\cellcolor[HTML]{FFFFFF}{\color[HTML]{000000} Cloudy}} &
  {\color[HTML]{000000} LPIPS$\downarrow$ ($\times$e-2)} &
  {\color[HTML]{000000} 0.897} &
  {\color[HTML]{000000} 1.997} &
  {\color[HTML]{000000} 0.772} &
  {\color[HTML]{000000} 1.142} &
  {\color[HTML]{000000} 0.641} &
  {\color[HTML]{000000} 0.918} &
  {\color[HTML]{000000} 0.879} &
  {\color[HTML]{000000} 4.680} &
  \multicolumn{1}{c|}{\cellcolor[HTML]{FFFFFF}{\color[HTML]{000000} {\ul 0.558}}} &
  {\color[HTML]{000000} \textbf{0.487}} \\ \hline
\cellcolor[HTML]{FFFFFF}{\color[HTML]{000000} } &
  {\color[HTML]{000000} fMAE$\downarrow$} &
  {\color[HTML]{000000} 0.171} &
  {\color[HTML]{000000} 0.122} &
  {\color[HTML]{000000} {\ul 0.085}} &
  {\color[HTML]{000000} 0.093} &
  {\color[HTML]{000000} \textbf{0.080}} &
  {\color[HTML]{000000} 0.098} &
  {\color[HTML]{000000} 0.182} &
  {\color[HTML]{000000} 0.136} &
  \multicolumn{1}{c|}{\cellcolor[HTML]{FFFFFF}{\color[HTML]{000000} 0.088}} &
  {\color[HTML]{000000} 0.094} \\
\cellcolor[HTML]{FFFFFF}{\color[HTML]{000000} } &
  {\color[HTML]{000000} fPSNR$\uparrow$} &
  {\color[HTML]{000000} 16.07} &
  {\color[HTML]{000000} 17.65} &
  {\color[HTML]{000000} {\ul 20.81}} &
  {\color[HTML]{000000} 20.21} &
  {\color[HTML]{000000} \textbf{21.41}} &
  {\color[HTML]{000000} 20.20} &
  {\color[HTML]{000000} 15.15} &
  {\color[HTML]{000000} 17.34} &
  \multicolumn{1}{c|}{\cellcolor[HTML]{FFFFFF}{\color[HTML]{000000} 20.16}} &
  {\color[HTML]{000000} 19.80} \\
\cellcolor[HTML]{FFFFFF}{\color[HTML]{000000} } &
  {\color[HTML]{000000} fSSIM$\uparrow$} &
  {\color[HTML]{000000} 0.701} &
  {\color[HTML]{000000} 0.736} &
  {\color[HTML]{000000} 0.819} &
  {\color[HTML]{000000} 0.821} &
  {\color[HTML]{000000} 0.818} &
  {\color[HTML]{000000} 0.791} &
  {\color[HTML]{000000} 0.690} &
  {\color[HTML]{000000} 0.746} &
  \multicolumn{1}{c|}{\cellcolor[HTML]{FFFFFF}{\color[HTML]{000000} {\ul 0.840}}} &
  {\color[HTML]{000000} \textbf{0.849}} \\
\multirow{-4}{*}{\cellcolor[HTML]{FFFFFF}{\color[HTML]{000000} Night}} &
  {\color[HTML]{000000} LPIPS$\downarrow$ ($\times$e-2)} &
  {\color[HTML]{000000} 2.078} &
  {\color[HTML]{000000} 2.264} &
  {\color[HTML]{000000} 1.291} &
  {\color[HTML]{000000} 1.777} &
  {\color[HTML]{000000} {\ul 1.127}} &
  {\color[HTML]{000000} 1.501} &
  {\color[HTML]{000000} 2.154} &
  {\color[HTML]{000000} 4.252} &
  \multicolumn{1}{c|}{\cellcolor[HTML]{FFFFFF}{\color[HTML]{000000} 1.146}} &
  {\color[HTML]{000000} \textbf{1.105}} \\ \hline
\cellcolor[HTML]{FFFFFF}{\color[HTML]{000000} } &
  {\color[HTML]{000000} fMAE$\downarrow$} &
  {\color[HTML]{000000} 0.093} &
  {\color[HTML]{000000} 0.109} &
  {\color[HTML]{000000} 0.072} &
  {\color[HTML]{000000} 0.075} &
  {\color[HTML]{000000} 0.065} &
  {\color[HTML]{000000} 0.073} &
  {\color[HTML]{000000} 0.088} &
  {\color[HTML]{000000} 0.105} &
  \multicolumn{1}{c|}{\cellcolor[HTML]{FFFFFF}{\color[HTML]{000000} {\ul 0.061}}} &
  {\color[HTML]{000000} \textbf{0.056}} \\
\cellcolor[HTML]{FFFFFF}{\color[HTML]{000000} } &
  {\color[HTML]{000000} fPSNR$\uparrow$} &
  {\color[HTML]{000000} 20.53} &
  {\color[HTML]{000000} 18.63} &
  {\color[HTML]{000000} 21.95} &
  {\color[HTML]{000000} 21.72} &
  {\color[HTML]{000000} 22.83} &
  {\color[HTML]{000000} 21.96} &
  {\color[HTML]{000000} 20.88} &
  {\color[HTML]{000000} 19.22} &
  \multicolumn{1}{c|}{\cellcolor[HTML]{FFFFFF}{\color[HTML]{000000} {\ul 23.21}}} &
  {\color[HTML]{000000} \textbf{23.86}} \\
\cellcolor[HTML]{FFFFFF}{\color[HTML]{000000} } &
  {\color[HTML]{000000} fSSIM$\uparrow$} &
  {\color[HTML]{000000} 0.840} &
  {\color[HTML]{000000} 0.810} &
  {\color[HTML]{000000} 0.859} &
  {\color[HTML]{000000} 0.876} &
  {\color[HTML]{000000} 0.868} &
  {\color[HTML]{000000} 0.841} &
  {\color[HTML]{000000} 0.848} &
  {\color[HTML]{000000} 0.803} &
  \multicolumn{1}{c|}{\cellcolor[HTML]{FFFFFF}{\color[HTML]{000000} {\ul 0.900}}} &
  {\color[HTML]{000000} \textbf{0.918}} \\
\multirow{-4}{*}{\cellcolor[HTML]{FFFFFF}{\color[HTML]{000000} All}} &
  {\color[HTML]{000000} LPIPS$\downarrow$ ($\times$e-2)} &
  {\color[HTML]{000000} 1.028} &
  {\color[HTML]{000000} 2.213} &
  {\color[HTML]{000000} 0.967} &
  {\color[HTML]{000000} 1.243} &
  {\color[HTML]{000000} 0.741} &
  {\color[HTML]{000000} 1.038} &
  {\color[HTML]{000000} 1.033} &
  {\color[HTML]{000000} 3.900} &
  \multicolumn{1}{c|}{\cellcolor[HTML]{FFFFFF}{\color[HTML]{000000} {\ul 0.693}}} &
  {\color[HTML]{000000} \textbf{0.596}} \\ \hline
{\color[HTML]{000000} \textbf{}} &
  {\color[HTML]{000000} Parameters$\downarrow$} &
  {\color[HTML]{000000} -} &
  {\color[HTML]{000000} -} &
  {\color[HTML]{000000} 54.756M} &
  {\color[HTML]{000000} 40.863M} &
  {\color[HTML]{000000} 54.763M} &
  {\color[HTML]{000000} 34.299M} &
  {\color[HTML]{000000} \textbf{2.744M}} &
  {\color[HTML]{000000} 44.900M} &
  \multicolumn{2}{c|}{\cellcolor[HTML]{FFFFFF}{\color[HTML]{000000} {\ul 10.403M} }} \\ \hline
\end{tabular}
}
\label{tab:sota}
\end{table*}

%% file: Tables/Illum_Resp.tex
\begin{table}
\centering
\caption{
Quantitative comparison of different illumination representations for image harmonization on the test set.
} % \caption
\resizebox{0.8\linewidth}{!}{ %< auto-adjusts font size to fill line
\begin{tabular}{@{}c|cccc@{}}
\toprule
 & fMAE$\downarrow$ & fPSNR$\uparrow$ & fSSIM$\uparrow$ & LPIPS$\downarrow$\\
\midrule
HDR illum. map &  0.097 & 19.91 & 0.858 & 0.017\\
Our illum. descriptor  & \textbf{0.061} & \textbf{23.21} & \textbf{0.900} & \textbf{0.007}\\
\bottomrule
\end{tabular}
} %< \resizebox
\label{tab:ablations}
\end{table}

%% file: Tables/user_study.tex
\begin{table}[ht]
\setlength{\tabcolsep}{2mm}
\centering
\caption{User study on real data.}
\resizebox{0.99\linewidth}{!}{

\begin{tabular}{l|cccc}
\hline
{\color[HTML]{000000} }                                     & {\color[HTML]{000000} Score (Q1)} & {\color[HTML]{000000} Score (Q2)} & {\color[HTML]{000000} Score (Q3)} & {\color[HTML]{000000} Overall Score} \\ \hline
Lalonde and Efros \cite{lalonde2007using} & 0.987                              & 1.084                              & 1.893                              & 1.322                               \\

DIH-GAN  \cite{bao2022deep}                                                  & 1.293                              & 1.466                              & 1.017                              & 1.259                               \\

DoveNet \cite{Cong2020DoveNet}                                                  & 1.670                              & 1.480                              & 1.883                              & 1.678                               \\
RainNet  \cite{ling2021region}                                                  & 1.636                              & 1.469                              & 1.720                              & 1.608                               \\
Guo et al.~\cite{guo2021iih}                                                  & 1.512                              & 1.477                              & 1.633                              & 1.541                               \\
D-HT~\cite{guo2021dht}                                                  & 1.656                              & 1.494                              & 1.901                              & 1.684                               \\
SCS-Co  \cite{hang2022scs}                                                  & 1.645                              &  1.473                              &  1.897                              &  1.672                               \\
CDTNet  \cite{cong2022high}                                                  &  1.749                              &  1.497                              &  1.912                              &  1.719                               \\

Ours                                                        & \textbf{2.051}                     & \textbf{1.915}                     & \textbf{1.981}                     & \textbf{1.982}                      \\ \hline
\end{tabular}
}
\label{tab:user-study}
\end{table}

%% file: Tables/AblationNRF.tex
\begin{table}[]
\centering
\caption{Ablation study on neural rendering framework.}
\resizebox{0.99\linewidth}{!}{ %< auto-adjusts font size to fill line
\begin{tabular}{@{}l|ccc@{}}
\toprule
& fPSNR$\uparrow$ & fSSIM$\uparrow$ & LPIPS$\downarrow$ ($\times$e-2)  \\ \midrule
Baseline NRF                    & 22.63 & 0.893 & 0.775 \\
Baseline NRF + $F_{a}$          & 23.25 & 0.897 & 0.735 \\
Baseline NRF + $F_{a}$ + $\hat{I}$  & 23.18 & 0.901 & 0.704 \\
Baseline NRF + $F_{a}$ + $\hat{I}$ + $B$ & 23.82 & 0.906 & 0.647 \\
Baseline NRF + $F_{a}$ + $\hat{I}$ + $B$ + SSIM loss & \textbf{23.86} & \textbf{0.918} & \textbf{0.596} \\ \bottomrule
\end{tabular}
}

\label{tab:ablation_NRF}
\end{table}

%% file: Tables/AblationK.tex
\begin{table}
\centering
\caption{
Effects of the number of shading bases ($K$).
} % \caption
%\resizebox{0.7\linewidth}{!}{ %< auto-adjusts font size to fill line
\begin{tabular}{@{}c|cccccc@{}}
\toprule
$K$   & 4     & 8     & 16    & 32    & 64    & 128             \\ \midrule
fPSNR$\uparrow$ & 29.44 & 30.81 & 31.21 & 31.35 & \textbf{31.42} & 31.41 \\
fSSIM$\uparrow$ & 0.919 & 0.931 & 0.935 & 0.936 & \textbf{0.937} & \textbf{0.937} \\ \bottomrule
\end{tabular}
%} %< \resizebox
\label{tab:ablation_k}
\end{table}

%% file: bare_jrnl_new_sample4.bbl
\begin{thebibliography}{10}
\providecommand{\url}[1]{#1}
\csname url@samestyle\endcsname
\providecommand{\newblock}{\relax}
\providecommand{\bibinfo}[2]{#2}
\providecommand{\BIBentrySTDinterwordspacing}{\spaceskip=0pt\relax}
\providecommand{\BIBentryALTinterwordstretchfactor}{4}
\providecommand{\BIBentryALTinterwordspacing}{\spaceskip=\fontdimen2\font plus
\BIBentryALTinterwordstretchfactor\fontdimen3\font minus
  \fontdimen4\font\relax}
\providecommand{\BIBforeignlanguage}[2]{{%
\expandafter\ifx\csname l@#1\endcsname\relax
\typeout{** WARNING: IEEEtran.bst: No hyphenation pattern has been}%
\typeout{** loaded for the language `#1'. Using the pattern for}%
\typeout{** the default language instead.}%
\else
\language=\csname l@#1\endcsname
\fi
#2}}
\providecommand{\BIBdecl}{\relax}
\BIBdecl

\bibitem{Cong2020DoveNet}
W.~Cong, J.~Zhang, L.~Niu, L.~Liu, Z.~Ling, W.~Li, and L.~Zhang, ``{DoveNet}:
  Deep image harmonization via domain verification,'' in \emph{CVPR}, 2020.

\bibitem{ling2021region}
J.~Ling, H.~Xue, L.~Song, R.~Xie, and X.~Gu, ``Region-aware adaptive instance
  normalization for image harmonization,'' in \emph{Proceedings of the IEEE/CVF
  Conference on Computer Vision and Pattern Recognition}, 2021, pp. 9361--9370.

\bibitem{lalonde2007using}
J.-F. Lalonde and A.~A. Efros, ``Using color compatibility for assessing image
  realism,'' in \emph{2007 IEEE 11th International Conference on Computer
  Vision}.\hskip 1em plus 0.5em minus 0.4em\relax IEEE, 2007, pp. 1--8.

\bibitem{xue2012understanding}
S.~Xue, A.~Agarwala, J.~Dorsey, and H.~Rushmeier, ``Understanding and improving
  the realism of image composites,'' \emph{ACM Transactions on graphics (TOG)},
  vol.~31, no.~4, pp. 1--10, 2012.

\bibitem{zhu2015learning}
J.-Y. Zhu, P.~Krahenbuhl, E.~Shechtman, and A.~A. Efros, ``Learning a
  discriminative model for the perception of realism in composite images,'' in
  \emph{Proceedings of the IEEE International Conference on Computer Vision},
  2015, pp. 3943--3951.

\bibitem{tsai2017deep}
Y.-H. Tsai, X.~Shen, Z.~Lin, K.~Sunkavalli, X.~Lu, and M.-H. Yang, ``Deep image
  harmonization,'' in \emph{Proceedings of the IEEE Conference on Computer
  Vision and Pattern Recognition}, 2017, pp. 3789--3797.

\bibitem{cong2021deep}
J.~Cao, W.~Cong, L.~Niu, J.~Zhang, and L.~Zhang, ``Deep image harmonization by
  bridging the reality gap,'' in \emph{33rd British Machine Vision Conference
  2022, {BMVC} 2022, London, UK, November 21-24, 2022}.\hskip 1em plus 0.5em
  minus 0.4em\relax {BMVA} Press, 2022.

\bibitem{guo2021iih}
Z.~Guo, H.~Zheng, Y.~Jiang, Z.~Gu, and B.~Zheng, ``Intrinsic image
  harmonization,'' in \emph{Proceedings of the IEEE/CVF Conference on Computer
  Vision and Pattern Recognition (CVPR)}, June 2021, pp. 16\,367--16\,376.

\bibitem{reinhard2001color}
E.~Reinhard, M.~Adhikhmin, B.~Gooch, and P.~Shirley, ``Color transfer between
  images,'' \emph{IEEE Computer graphics and applications}, vol.~21, no.~5, pp.
  34--41, 2001.

\bibitem{xiao2006color}
X.~Xiao and L.~Ma, ``Color transfer in correlated color space,'' in
  \emph{Proceedings of the 2006 ACM international conference on Virtual reality
  continuum and its applications}, 2006, pp. 305--309.

\bibitem{pitie2007automated}
F.~Piti{\'e}, A.~C. Kokaram, and R.~Dahyot, ``Automated colour grading using
  colour distribution transfer,'' \emph{Computer Vision and Image
  Understanding}, vol. 107, no. 1-2, pp. 123--137, 2007.

\bibitem{fecker2008histogram}
U.~Fecker, M.~Barkowsky, and A.~Kaup, ``Histogram-based prefiltering for
  luminance and chrominance compensation of multiview video,'' \emph{IEEE
  Transactions on Circuits and Systems for Video Technology}, vol.~18, no.~9,
  pp. 1258--1267, 2008.

\bibitem{debevec2006image}
P.~Debevec, ``Image-based lighting,'' in \emph{ACM SIGGRAPH 2006 Courses},
  2006, pp. 4--es.

\bibitem{bao2022deep}
Z.~Bao, C.~Long, G.~Fu, D.~Liu, Y.~Li, J.~Wu, and C.~Xiao, ``Deep image-based
  illumination harmonization,'' in \emph{Proceedings of the IEEE/CVF Conference
  on Computer Vision and Pattern Recognition}, 2022, pp. 18\,542--18\,551.

\bibitem{Debevec1997HDR}
\BIBentryALTinterwordspacing
P.~E. Debevec and J.~Malik, ``Recovering high dynamic range radiance maps from
  photographs,'' in \emph{Proceedings of the 24th Annual Conference on Computer
  Graphics and Interactive Techniques}, ser. SIGGRAPH '97.\hskip 1em plus 0.5em
  minus 0.4em\relax USA: ACM Press/Addison-Wesley Publishing Co., 1997, p.
  369–378. [Online]. Available: \url{https://doi.org/10.1145/258734.258884}
\BIBentrySTDinterwordspacing

\bibitem{ramamoorthi2001efficient}
R.~Ramamoorthi and P.~Hanrahan, ``An efficient representation for irradiance
  environment maps,'' in \emph{Proceedings of the 28th annual conference on
  Computer graphics and interactive techniques}, 2001, pp. 497--500.

\bibitem{perez2003poisson}
P.~P{\'e}rez, M.~Gangnet, and A.~Blake, ``Poisson image editing,'' in \emph{ACM
  SIGGRAPH 2003 Papers}, 2003, pp. 313--318.

\bibitem{pitie2005ndimensional}
F.~Pitie, A.~C. Kokaram, and R.~Dahyot, ``N-dimensional probability density
  function transfer and its application to color transfer,'' in \emph{Tenth
  IEEE International Conference on Computer Vision (ICCV'05) Volume 1},
  vol.~2.\hskip 1em plus 0.5em minus 0.4em\relax IEEE, 2005, pp. 1434--1439.

\bibitem{cohen2006color}
D.~Cohen-Or, O.~Sorkine, R.~Gal, T.~Leyvand, and Y.-Q. Xu, ``Color
  harmonization,'' in \emph{ACM SIGGRAPH 2006 Papers}, 2006, pp. 624--630.

\bibitem{Jia2006Pasting}
J.~Jia, J.~Sun, C.-K. Tang, and H.-Y. Shum, ``Drag-and-drop pasting,''
  \emph{ACM Transactions on Graphics (SIGGRAPH)}, 2006.

\bibitem{sunkavalli2010multi}
K.~Sunkavalli, M.~K. Johnson, W.~Matusik, and H.~Pfister, ``Multi-scale image
  harmonization,'' \emph{ACM Transactions on Graphics (TOG)}, vol.~29, no.~4,
  pp. 1--10, 2010.

\bibitem{song2020illumination}
S.~Song, F.~Zhong, X.~Qin, and C.~Tu, ``Illumination harmonization with gray
  mean scale,'' in \emph{Computer Graphics International Conference}.\hskip 1em
  plus 0.5em minus 0.4em\relax Springer, 2020, pp. 193--205.

\bibitem{sofiiuk2021foreground}
K.~Sofiiuk, P.~Popenova, and A.~Konushin, ``Foreground-aware semantic
  representations for image harmonization,'' in \emph{Proceedings of the
  IEEE/CVF Winter Conference on Applications of Computer Vision}, 2021, pp.
  1620--1629.

\bibitem{jiang2021ssh}
Y.~Jiang, H.~Zhang, J.~Zhang, Y.~Wang, Z.~Lin, K.~Sunkavalli, S.~Chen,
  S.~Amirghodsi, S.~Kong, and Z.~Wang, ``{SSH}: A self-supervised framework for
  image harmonization,'' in \emph{Proceedings of the IEEE/CVF International
  Conference on Computer Vision}, 2021, pp. 4832--4841.

\bibitem{cong2022high}
W.~Cong, X.~Tao, L.~Niu, J.~Liang, X.~Gao, Q.~Sun, and L.~Zhang,
  ``High-resolution image harmonization via collaborative dual
  transformations,'' in \emph{Proceedings of the IEEE/CVF Conference on
  Computer Vision and Pattern Recognition}, 2022, pp. 18\,470--18\,479.

\bibitem{cong2021bargainnet}
W.~Cong, L.~Niu, J.~Zhang, J.~Liang, and L.~Zhang, ``{BargainNet}:
  Background-guided domain translation for image harmonization,'' in \emph{2021
  IEEE International Conference on Multimedia and Expo (ICME)}.\hskip 1em plus
  0.5em minus 0.4em\relax IEEE, 2021, pp. 1--6.

\bibitem{hang2022scs}
Y.~Hang, B.~Xia, W.~Yang, and Q.~Liao, ``{SCS-Co}: Self-consistent style
  contrastive learning for image harmonization,'' in \emph{Proceedings of the
  IEEE/CVF Conference on Computer Vision and Pattern Recognition}, 2022, pp.
  19\,710--19\,719.

\bibitem{cun2020improving}
X.~Cun and C.-M. Pun, ``Improving the harmony of the composite image by
  spatial-separated attention module,'' \emph{IEEE Transactions on Image
  Processing}, vol.~29, pp. 4759--4771, 2020.

\bibitem{hao2020image}
G.~Hao, S.~Iizuka, and K.~Fukui, ``Image harmonization with attention-based
  deep feature modulation.'' in \emph{BMVC}, 2020.

\bibitem{guo2021dht}
Z.~Guo, D.~Guo, H.~Zheng, Z.~Gu, B.~Zheng, and J.~Dong, ``Image harmonization
  with transformer,'' in \emph{Proceedings of the IEEE/CVF International
  Conference on Computer Vision}, 2021, pp. 14\,870--14\,879.

\bibitem{debevec2000acquiring}
P.~Debevec, T.~Hawkins, C.~Tchou, H.-P. Duiker, W.~Sarokin, and M.~Sagar,
  ``Acquiring the reflectance field of a human face,'' in \emph{Proceedings of
  the 27th annual conference on Computer graphics and interactive techniques},
  2000, pp. 145--156.

\bibitem{xu2018deep}
Z.~Xu, K.~Sunkavalli, S.~Hadap, and R.~Ramamoorthi, ``Deep image-based
  relighting from optimal sparse samples,'' \emph{ACM Transactions on Graphics
  (TOG)}, vol.~37, no.~4, pp. 1--13, 2018.

\bibitem{meka2019deep}
A.~Meka, C.~Haene, R.~Pandey, M.~Zollh{\"o}fer, S.~Fanello, G.~Fyffe,
  A.~Kowdle, X.~Yu, J.~Busch, J.~Dourgarian \emph{et~al.}, ``Deep reflectance
  fields: high-quality facial reflectance field inference from color gradient
  illumination,'' \emph{ACM Transactions on Graphics (TOG)}, vol.~38, no.~4,
  pp. 1--12, 2019.

\bibitem{zhou2019deep}
H.~Zhou, S.~Hadap, K.~Sunkavalli, and D.~W. Jacobs, ``Deep single-image
  portrait relighting,'' in \emph{Proceedings of the IEEE International
  Conference on Computer Vision}, 2019, pp. 7194--7202.

\bibitem{sun2019single}
T.~Sun, J.~T. Barron, Y.-T. Tsai, Z.~Xu, X.~Yu, G.~Fyffe, C.~Rhemann, J.~Busch,
  P.~E. Debevec, and R.~Ramamoorthi, ``Single image portrait relighting.''
  \emph{ACM Trans. Graph.}, vol.~38, no.~4, pp. 79--1, 2019.

\bibitem{kanamori2018relighting}
Y.~Kanamori and Y.~Endo, ``Relighting humans: occlusion-aware inverse rendering
  for full-body human images,'' \emph{ACM Transactions on Graphics (TOG)},
  vol.~37, no.~6, pp. 1--11, 2018.

\bibitem{wang2020single}
Z.~Wang, X.~Yu, M.~Lu, Q.~Wang, C.~Qian, and F.~Xu, ``Single image portrait
  relighting via explicit multiple reflectance channel modeling,'' \emph{ACM
  Transactions on Graphics (TOG)}, vol.~39, no.~6, pp. 1--13, 2020.

\bibitem{sang2020single}
S.~Sang and M.~Chandraker, ``Single-shot neural relighting and {SVBRDF}
  estimation,'' in \emph{European Conference on Computer Vision}.\hskip 1em
  plus 0.5em minus 0.4em\relax Springer, 2020, pp. 85--101.

\bibitem{lagunas2021single}
M.~Lagunas, X.~Sun, J.~Yang, R.~Villegas, J.~Zhang, Z.~Shu, B.~Masia, and
  D.~Gutierrez, ``Single-image full-body human relighting,'' in
  \emph{Eurographics Symposium on Rendering (EGSR)}.\hskip 1em plus 0.5em minus
  0.4em\relax The Eurographics Association, 2021.

\bibitem{yu2020self}
Y.~Yu, A.~Meka, M.~Elgharib, H.-P. Seidel, C.~Theobalt, and W.~A. Smith,
  ``Self-supervised outdoor scene relighting,'' in \emph{European Conference on
  Computer Vision}.\hskip 1em plus 0.5em minus 0.4em\relax Springer, 2020, pp.
  84--101.

\bibitem{3dpeople}
``{3D} {People},'' \url{https://3dpeople.com}.

\bibitem{polyHaven}
``{Poly} {Haven},'' \url{https://polyhaven.com/hdris}.

\bibitem{hdrmaps}
``{HDR} {MAPS},'' \url{https://hdrmaps.com/}.

\bibitem{blender}
\BIBentryALTinterwordspacing
B.~O. Community, \emph{Blender - a 3D modelling and rendering package}, Blender
  Foundation, Stichting Blender Foundation, Amsterdam, 2018. [Online].
  Available: \url{http://www.blender.org}
\BIBentrySTDinterwordspacing

\bibitem{belhumeur1998set}
P.~N. Belhumeur and D.~J. Kriegman, ``What is the set of images of an object
  under all possible illumination conditions?'' \emph{International journal of
  computer vision}, vol.~28, no.~3, pp. 245--260, 1998.

\bibitem{ronneberger2015u}
O.~Ronneberger, P.~Fischer, and T.~Brox, ``{U-Net}: Convolutional networks for
  biomedical image segmentation,'' in \emph{International Conference on Medical
  image computing and computer-assisted intervention}.\hskip 1em plus 0.5em
  minus 0.4em\relax Springer, 2015, pp. 234--241.

\bibitem{zhang2018residual}
Y.~Zhang, Y.~Tian, Y.~Kong, B.~Zhong, and Y.~Fu, ``Residual dense network for
  image super-resolution,'' in \emph{Proceedings of the IEEE conference on
  computer vision and pattern recognition}, 2018, pp. 2472--2481.

\bibitem{vaswani2017attention}
A.~Vaswani, N.~Shazeer, N.~Parmar, J.~Uszkoreit, L.~Jones, A.~N. Gomez,
  {\L}.~Kaiser, and I.~Polosukhin, ``Attention is all you need,'' in
  \emph{Advances in neural information processing systems}, 2017, pp.
  5998--6008.

\bibitem{zhao2016loss}
H.~Zhao, O.~Gallo, I.~Frosio, and J.~Kautz, ``Loss functions for image
  restoration with neural networks,'' \emph{IEEE Transactions on computational
  imaging}, vol.~3, no.~1, pp. 47--57, 2016.

\bibitem{wang2004image}
Z.~Wang, A.~C. Bovik, H.~R. Sheikh, and E.~P. Simoncelli, ``Image quality
  assessment: from error visibility to structural similarity,'' \emph{IEEE
  transactions on image processing}, vol.~13, no.~4, pp. 600--612, 2004.

\bibitem{zhang2018perceptual}
R.~Zhang, P.~Isola, A.~A. Efros, E.~Shechtman, and O.~Wang, ``The unreasonable
  effectiveness of deep features as a perceptual metric,'' in \emph{CVPR},
  2018.

\bibitem{taobao}
``{Taobao},'' \url{https://www.taobao.com/}.

\bibitem{ashikmin2000microfacet}
M.~Ashikmin, S.~Premo{\v{z}}e, and P.~Shirley, ``A microfacet-based {BRDF}
  generator,'' in \emph{Proceedings of the 27th annual conference on Computer
  graphics and interactive techniques}, 2000, pp. 65--74.

\end{thebibliography}
